\documentclass[10pt,twocolumn,letterpaper]{article}

\usepackage{cvpr}              

\usepackage{graphicx}
\usepackage{amsmath}
\usepackage{amssymb}
\usepackage{booktabs}
\usepackage{bbm}
\usepackage{diagbox}

\usepackage[pagebackref,breaklinks,colorlinks]{hyperref}

\usepackage{makecell}
\usepackage[para]{footmisc}

\usepackage[capitalize]{cleveref}
\crefname{section}{Sec.}{Secs.}
\Crefname{section}{Section}{Sections}
\Crefname{table}{Table}{Tables}
\crefname{table}{Tab.}{Tabs.}

\def\cvprPaperID{2545} 
\def\confName{CVPR}
\def\confYear{2023}

\usepackage{xcolor}
\newcount\Comments  
\newcommand{\kibitz}[2]{\ifnum\Comments=0\textcolor{#1}{#2}\fi}

\begin{document}

\title{Physically Realizable Natural-Looking Clothing Textures Evade Person Detectors via 3D Modeling}

\author{Zhanhao Hu$^{1}$\thanks{Equal contribution.} \ \ Wenda Chu$^{2, 1}$\footnotemark[1] \ \ Xiaopei Zhu$^{3, 1}$ \ \ Hui Zhang$^{4}$ \ \ Bo Zhang$^{1}$ \ \ Xiaolin Hu$^{1,5,6}$\thanks{Corresponding author.}\\
$^{1}$Department of Computer Science and Technology, Tsinghua University, Beijing, China \\
$^{2}$Institute for Interdisciplinary Information Sciences, Tsinghua University, Beijing, China \\
$^{3}$School of Integrated Circuits, Tsinghua University, Beijing, China \\
$^{4}$Beijing Institute of Fashion Technology, Beijing, China\\
$^{5}$IDG/McGovern Institute for Brain Research, THBI, Tsinghua University, Beijing, China \\
$^{6}$Chinese Institute for Brain Research (CIBR), Beijing, China\\
\tt\small \{huzhanha17, chuwd19, zxp18\}@mails.tsinghua.edu.cn \\ \tt\small fzyzhh@bift.edu.cn, \tt\small \{dcszb, xlhu\}@mail.tsinghua.edu.cn}

\maketitle

\begin{abstract}
   Recent works have proposed to craft adversarial clothes for evading person detectors, while they are either only effective at limited viewing angles or very conspicuous to humans. We aim to craft adversarial texture for clothes based on 3D modeling, an idea that has been used to craft rigid adversarial objects such as a 3D-printed turtle. Unlike rigid objects,  humans and clothes are non-rigid, leading to difficulties in physical realization. In order to craft natural-looking adversarial clothes that can evade person detectors at multiple viewing angles, we propose adversarial camouflage textures (AdvCaT) that resemble one kind of the typical textures of daily clothes, camouflage textures. We leverage the Voronoi diagram and Gumbel-softmax trick to parameterize the camouflage textures and optimize the parameters via 3D modeling. Moreover, we propose an efficient augmentation pipeline on 3D meshes combining topologically plausible projection (TopoProj) and Thin Plate Spline (TPS) to narrow the gap between digital and real-world objects. We printed the developed 3D texture pieces on fabric materials and tailored them into T-shirts and trousers. Experiments show high attack success rates of these clothes against multiple detectors.
   
\end{abstract}

\section{Introduction}

\begin{figure}[ht]
    \centering
    \includegraphics[width=0.9\linewidth]{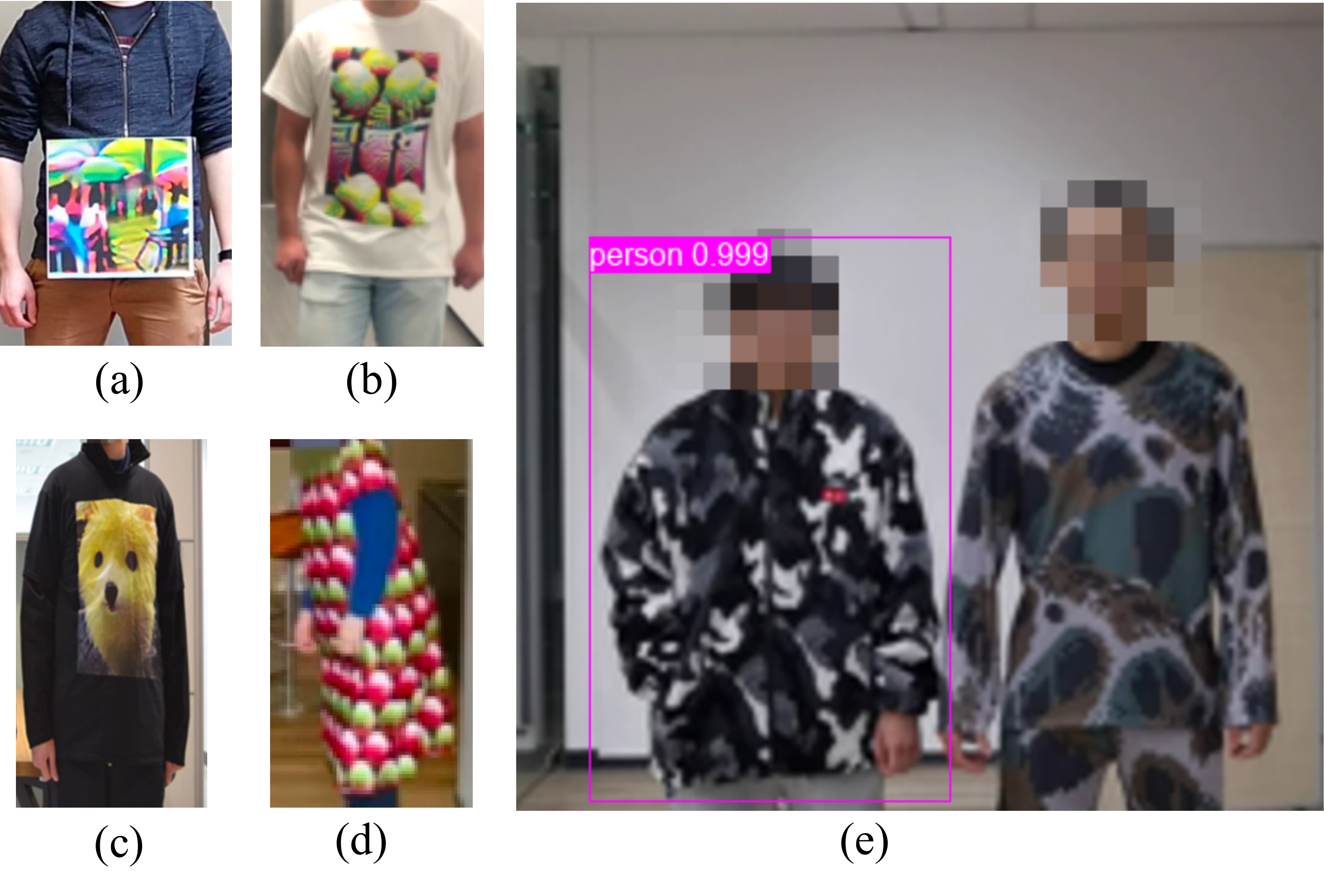}
    \caption{Visualization of several adversarial clothes. (a) Adversarial patch~\cite{thys2019fooling}. (b) Adversarial T-shirt~\cite{xu2020adversarial}. (c) Naturalistic patch~\cite{hu2021naturalistic}. (d) Adversarial Texture~\cite{hu2022adversarial}. (e) Left: daily camouflage texture; Right: our adversarial camouflage texture.}
    \label{fig:front}
\end{figure}
Deep Neural Networks(DNNs) have been widely used in many real-world systems such as face recognition and object detection \cite{Taigman2014DeepFace,redmon2018yolov3,ren2016faster,Zhu2021Deformable}. However, it is well known that DNNs are vulnerable to adversarial examples~\cite{goodfellow2014explaining,szegedy2013intriguing}. Adversarial examples can be crafted by adding small perturbations to the clean inputs, rendering the DNNs' outputs incorrect. Such vulnerabilities could result in severe safety problems when deploying DNN-based systems. This has become a hot research topic recently \cite{kurakin2016adversarial,papernot2016the,nguyen2015deep,moosavi2016deepfool,carlini2017towards,dong2018boosting}.

\begin{figure*}[t]
    \centering
    \includegraphics[width=0.85\textwidth]{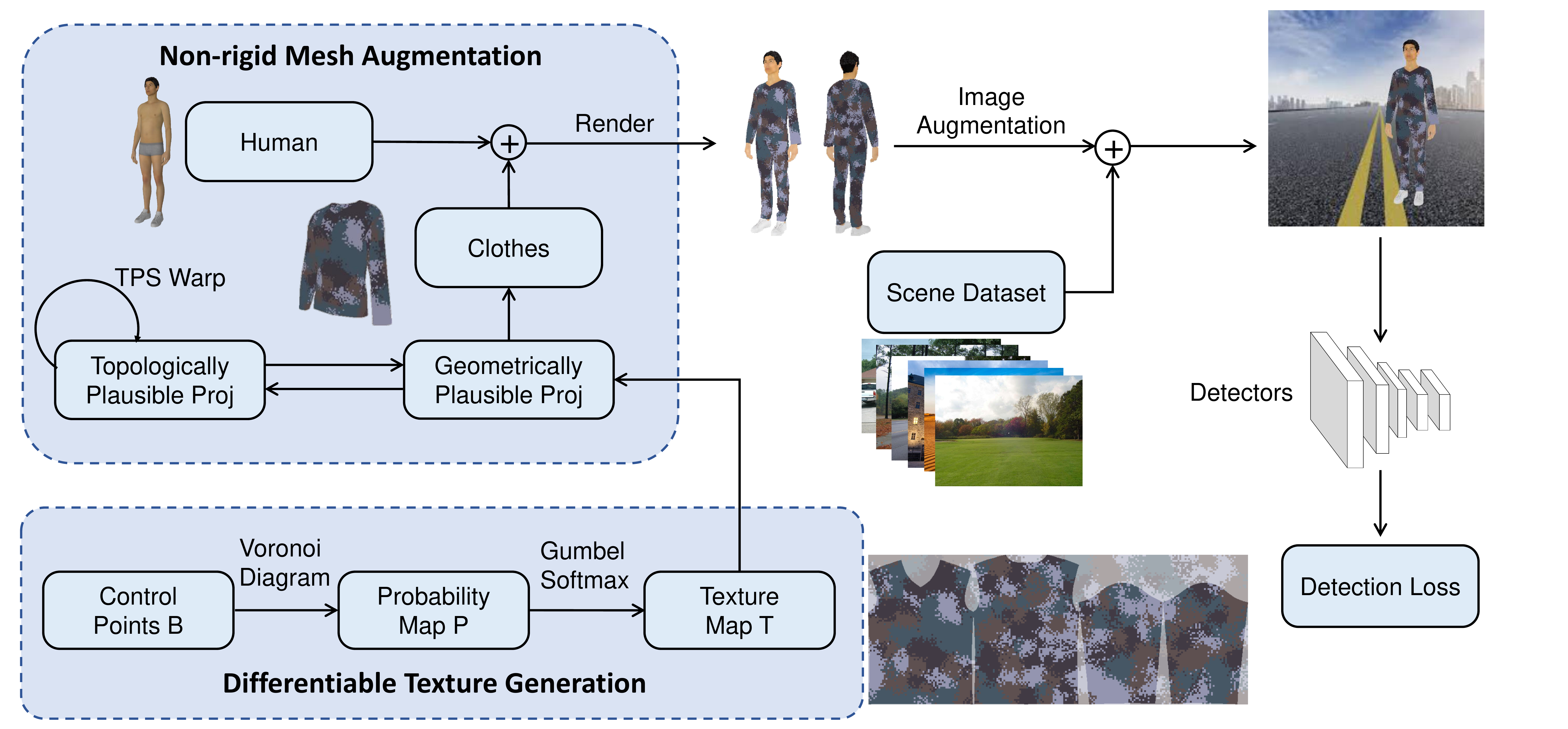}
    \caption{The training pipeline of the adversarial camouflage textures.}
    \label{fig:pipeline}
\end{figure*}

Adversarial examples were first identified in the digital world. However, adversarial examples also exist in the physical world, posing more risks in real-world scenarios. Recently, many works \cite{sharif2016accessorize,evtimov2017robust,athalye2018synthesizing,wang2020can,duan2020adversarial,wang2021dual,Zhu2021Fooling,zhong2022shadows,wang2022fca,zhu2022infrared,duan2022learning} have designed \emph{physical adversarial examples} to deceive DNNs in the real world. Among them, hiding persons\cite{thys2019fooling,xu2020adversarial,huang2020universal,wu2020making,hu2021naturalistic,hu2022adversarial} from DNN-based object detectors is especially challenging due to the difficulties of modeling non-rigid object surfaces (i.e., clothes). Most works \cite{thys2019fooling,xu2020adversarial,huang2020universal,wu2020making,hu2021naturalistic} print adversarial patches on the front side of clothes to hide people from being detected. We call them \emph{patch-based adversarial examples}. These patches are usually conspicuous to humans, making the clothes look strange and easily noticeable by human observers. Efforts have been put on making the adversarial patches more natural-looking~\cite{hu2021naturalistic,duan2020adversarial,wang2021dual}. However, these patch-based adversarial clothes can only attack object detectors at a narrow range of viewing angles (i.e., when the camera faces the front of the person). To attack the detector at a wider range of viewing angles, one may print the adversarial patches everywhere on the clothes, which would make the clothes unnatural-looking again. For example, a dog-head-like patch on the front of a T-shirt is natural, but putting this patch everywhere on the T-shirt would make the T-shirt look weird. 

Another way to craft physical adversarial examples is to design the textures on the surface of the target objects \cite{athalye2018synthesizing,hu2022adversarial,zhang2019camou,wang2022fca,duan2022learning}, i.e., crafting \emph{texture-base adversarial examples}. Unlike patch-based adversarial examples, texture-based ones are usually adversarially effective at multiple viewing angles. They are mostly optimized via 3D modeling or using clone networks, and printed on the surface of rigid objects such as turtles\cite{athalye2018synthesizing} and cars\cite{zhang2019camou,wang2022fca,duan2022learning}.
However, it is much harder to realize the 3D textures of non-rigid objects like humans and clothes  in the physical world while maintaining their adversarial effectiveness, since there is a huge gap between a 3D human model and a real-world person.
To circumvent this difficulty, Hu et al. \cite{hu2022adversarial} propose to craft texture-based adversarial clothes by extending patches into textures with repetitive patterns, which does not require 3D modeling. However, their textures are very conspicuous to humans, and obtaining natural-looking textures can be difficult under the constraint of repetitive patterns.

In this paper, we propose a 3D modeling pipeline to produce natural-looking adversarial clothes that are physically realizable and can hide people at multiple viewing angles. Specifically, we craft adversarial camouflage texture (AdvCaT) patterns and apply them on clothes. We choose camouflage texture patterns mainly because they are typical texture patterns widely used in daily clothes, therefore making the clothes more natural-looking
In order to make the texture patterns more generalizable when applied to deformed and unseen 3D models, we propose a novel 3D augmentation method combining topologically plausible projection (TopoProj) and thin plate spline (TPS)~\cite{bookstein1989principal,donato2002approximate,xu2020adversarial,tang2019augmentation} for non-rigid objects such as  clothes. 



We optimized several AdvCaT patterns to attack widely used detection models, including YOLOv3~\cite{redmon2018yolov3}, Faster RCNN~\cite{ren2016faster}, and deformable DETR~\cite{Zhu2021Deformable}, and applied the texture patterns on clothes in the physical world. See \cref{fig:front} for the visualization of our adversarial clothes compared with others. Experiments showed that our adversarial clothes could evade different detectors at multiple viewing angles. A subjective test experiment indicated that the naturalness score of our adversarial clothes covered with AdvCaT is significantly higher than other adversarial clothes and close to daily clothes.

\section{Related Work}
Early works~\cite{szegedy2013intriguing,goodfellow2014explaining,kurakin2016adversarial,carlini2017towards}  found that adversarial examples crafted by adding small digital adversarial perturbations on the clean inputs can mislead the DNNs. Some adversarial examples can also be crafted in the physical world to attack different DNN models, including image classification models~\cite{sharif2016accessorize,athalye2018synthesizing,brown2017adversarial,evtimov2017robust,zhong2022shadows} and detection models~\cite{chen2018shapeshifter,song2018physical,lee2019physical,thys2019fooling,huang2020universal,xu2020adversarial,wu2020making,wang2020can,hu2021naturalistic, Zhu2021Fooling}. Among these works, patch-based and texture-based attacks are typical ways to craft physical adversarial examples.

Patch-based attacks \cite{thys2019fooling,xu2020adversarial,huang2020universal,wu2020making,hu2021naturalistic, sharif2016accessorize, brown2017adversarial, evtimov2017robust,athalye2018synthesizing,zhang2019camou,wang2022fca,duan2022learning, huang2020universal} usually optimize patches and put them on the target objects, and therefore can only work at a narrow range of viewing angles. These works produce different adversarial objects, including glasses frames~\cite{sharif2016accessorize}, road signs~\cite{evtimov2017robust, gnanasambandam2021optical}, cars~\cite{zhang2019camou,wang2022fca, wang2021dual} and clothes~\cite{thys2019fooling,xu2020adversarial,huang2020universal,wu2020making,hu2021naturalistic,hu2022adversarial, huang2020universal}. Among them, hiding persons from object detectors is especially challenging since the adversarial patches on the clothes can be heavily deformed due to their non-rigidity ~\cite{xu2020adversarial}. On the other hand, these adversarial patches are usually conspicuous to humans. To this end, Duan et al.~\cite{duan2020adversarial} and Wang et al.~\cite{wang2021dual} introduce additional losses to make the adversarial patches less conspicuous. Hu et al. \cite{hu2021naturalistic} propose to produce more natural-looking patches with GANs~\cite{brock2019large,karras2020analyzing}.

Texture-based attacks~\cite{athalye2018synthesizing,hu2022adversarial,zhang2019camou,wang2022fca,zhu2022infrared}, on the other hand, optimize the textures on the surface of the target objects to craft physical adversarial examples. Covered with adversarial textures, the object usually can deceive DNNs at multiple viewing angles. These works mainly use 3D modeling or clone networks to optimize textures for rigid objects. Athalye et al. \cite{athalye2018synthesizing} introduce the Expectation over Transformation (EoT) method and produce an adversarial 3D-printed turtle. Zhang et al. \cite{zhang2019camou} and Wang et al. \cite{wang2022fca} design vehicle camouflage for multi-view adversarial attacks. Hu et al. \cite{hu2022adversarial} propose adversarial textures with repetitive structures for non-rigid clothes.

\section{Adversarial Camouflage Texture Patterns Generation}
In this section, we present the pipeline of generating adversarial camouflage texture (AdvCaT) that can be applied to clothes. As shown in \cref{fig:pipeline}, we adopt 3D meshes to model humans and clothes and define the surface of the clothes according to their 2D texture maps with UV coordinates. We propose two critical techniques to optimize adversarial camouflage texture clothes. The first is to parameterize the camouflage textures on the 2D texture maps with Voronoi diagram~\cite{voronoi} and Gumbel softmax trick~\cite{maddison2014sampling,gumbel2017}. The second is to apply a realistic deformation on the 3D meshes with the topologically plausible projection (TopoProj). 
We render the foreground photos of a 3D person wearing a T-shirt and a trouser using a differential renderer~\cite{ravi2020pytorch3d}. The foreground photos are synthesized with background images sampled from a scene dataset. Finally, we feed the synthesized photos into the victim detector and minimize an adversarial loss to optimize the parameters of camouflage patterns.

In what follows, we first introduce the differential generation of AdvCaT. Next, we present the novel 3D deformation which can be used to augment the meshes during training to boost the generalizability of the optimized textures. Finally, we elaborate the loss functions for optimization.



\begin{figure}[t]
    \centering
    \begin{subfigure}[t]{0.15\textwidth}
    \centering
    \includegraphics[width=\textwidth]{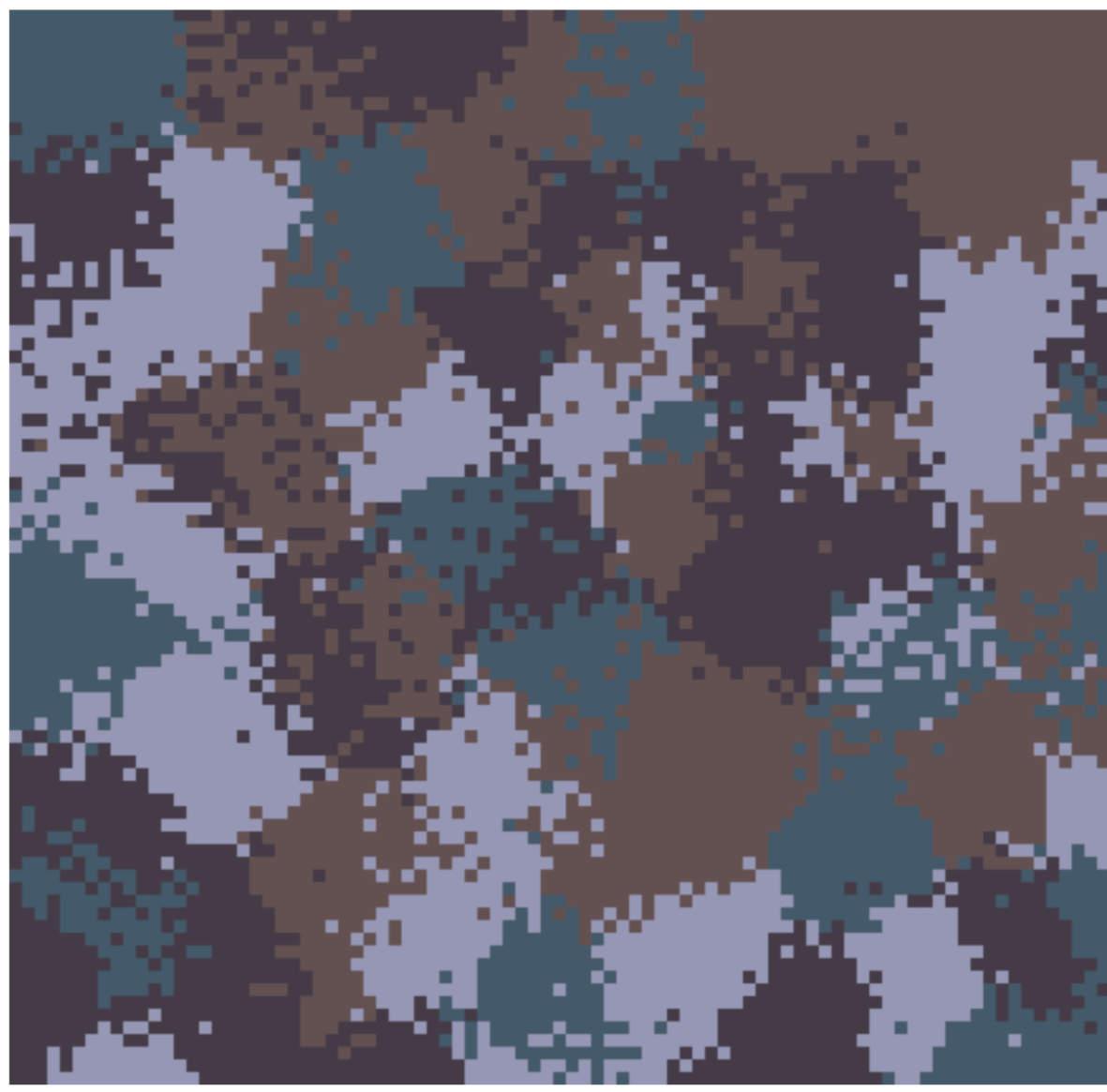}
    \caption{}
    \end{subfigure}
    \begin{subfigure}[t]{0.15\textwidth}
    \centering
    \includegraphics[width=\textwidth]{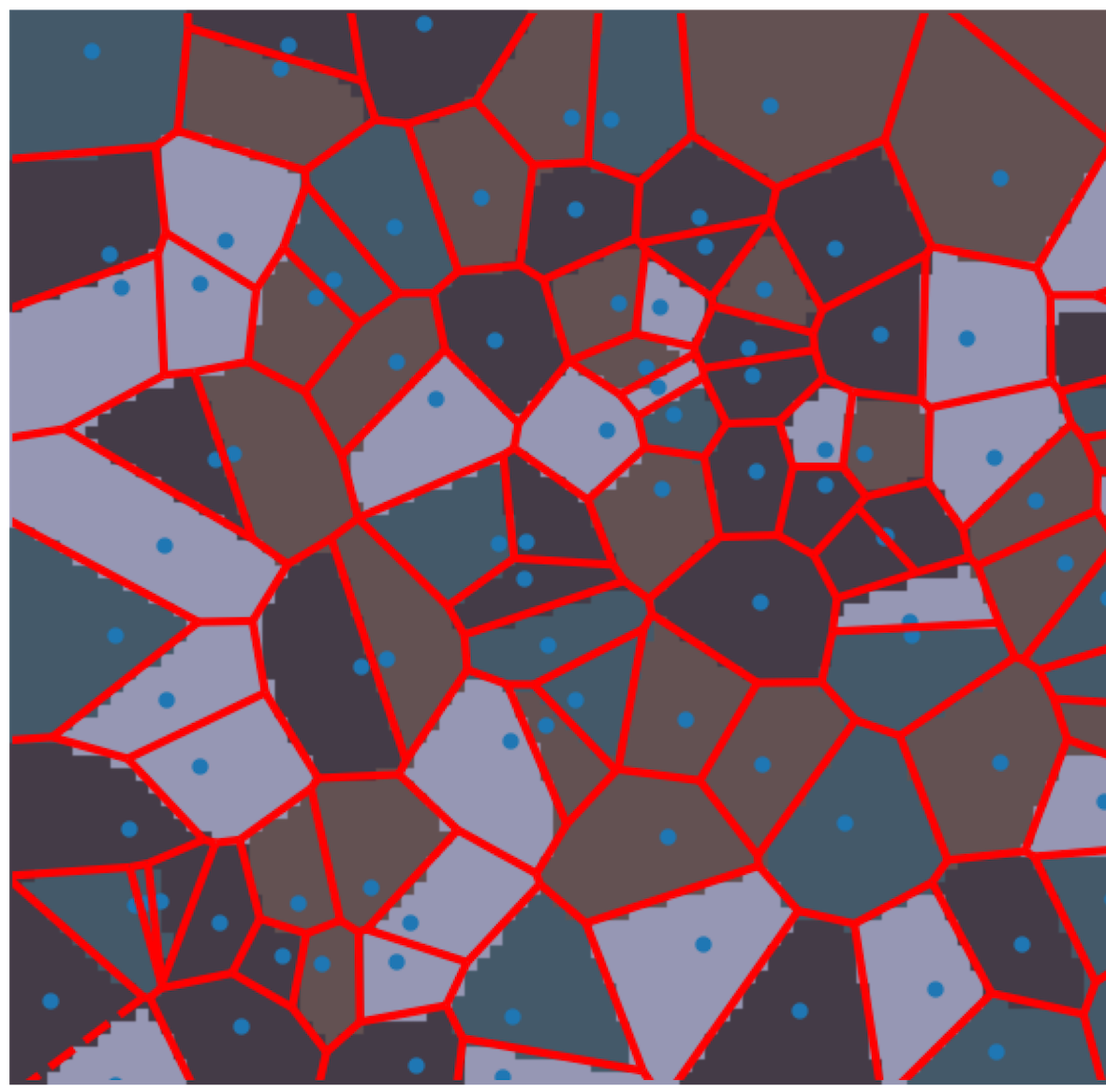}
    \caption{}
    \end{subfigure}
    \begin{subfigure}[t]{0.15\textwidth}
    \centering
    \includegraphics[width=\textwidth]{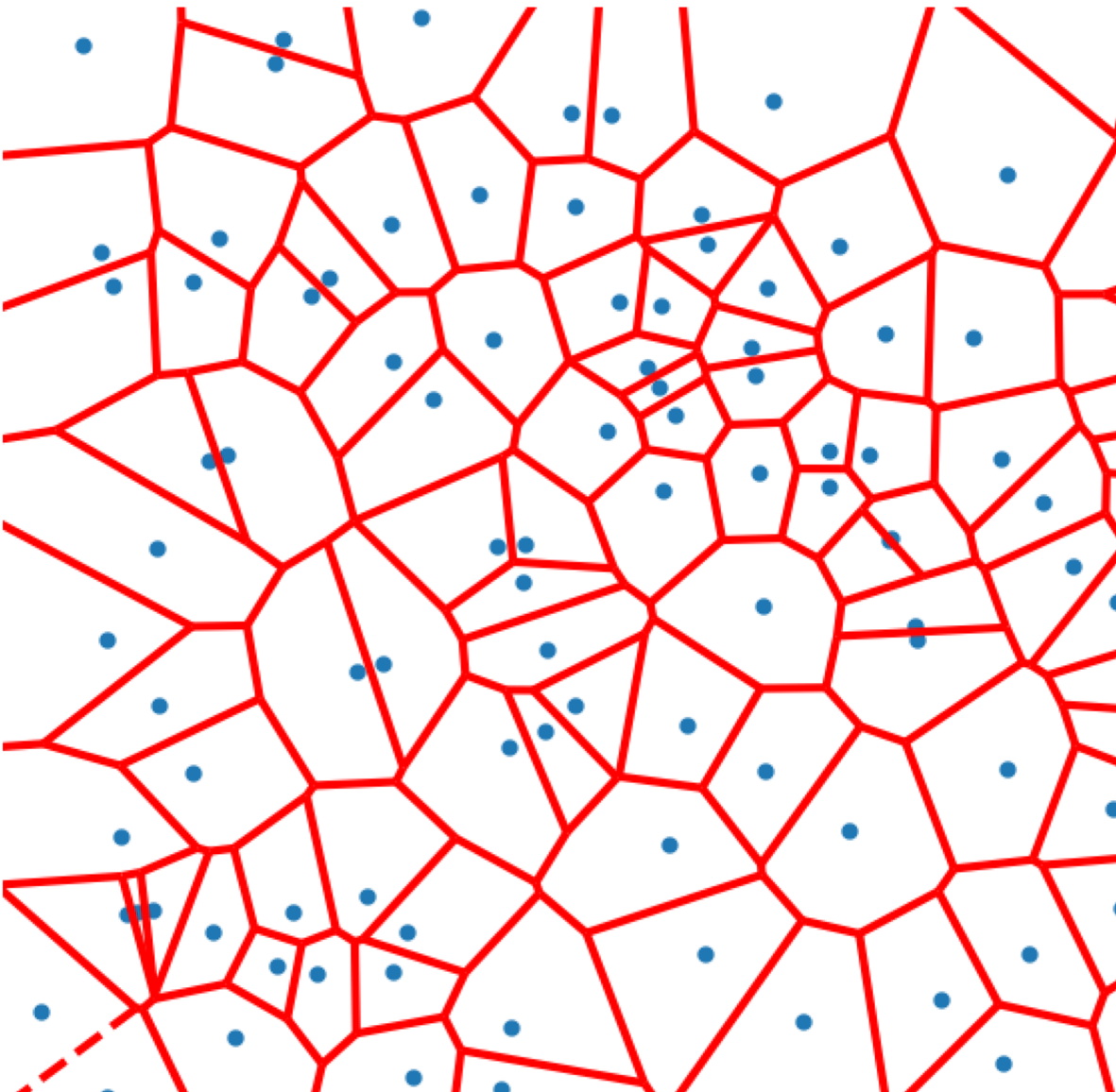}
    \caption{}
    \end{subfigure}
    \caption{(a) Camouflage texture. (b) Color cluster regions. The region of each color cluster can be approximately represented by polygons. (c) Voronoi diagram. The blue points are the control points, and the red lines are the boundaries of the regions.}
    \label{fig:VG}
\end{figure}

\subsection{Differentiable Generation of Camouflage Textures}
\label{subsection:camo}

Camouflage patterns are originally designed for concealing people in the wild and have now become typical textures of ordinary clothes. There are a few kinds of common camouflage patterns. Among them, we choose to imitate a specific type called digital camouflage patterns that consists of small rectangular pixels. As shown in ~\cref{fig:VG}a, digital camouflage patterns are typically irregular in shape and consist of a limited number of colors.

We noticed that the pixels of the camouflage patterns are locally
aggregated as clusters, each of which approximately covers
a polygon region. See ~\cref{fig:VG}b for illustration. Inspired by the polygon generation ability of Voronoi diagram~\cite{voronoi}, we use a soft version of Voronoi diagram to generate the cluster regions of the
camouflage pixels.



\vskip 5pt
\noindent\textbf{Polygon generation with Voronoi diagram.} 
A Voronoi diagram is a partition of a plane into multiple regions~\cite{voronoi}. Each region is controlled by a point, consisting of all the pixels closer to the corresponding control point than to any other point (see \cref{fig:VG}c). In this way, the locations and shapes of the polygons can be parameterized by the coordinates of the control points.

However, the locations and shapes of the polygons are not differentiable with respect to the coordinates of the control points if we directly apply this rule. Therefore, we define a soft version of Voronoi diagram by introducing probabilities for each pixel. Suppose the texture map only consists of several discrete colors in a color set $\mathcal{C}= \{c_i=(R_i,G_i,B_i)| i=1,\dots, N_C\}$. Then, $N_P$ independent control points are assigned to each color, with coordinates $\{b_{ij}\in \mathbb{R}^2, i=1,2,\dots,N_C, j= 1,2,\dots, N_P\}$. For each pixel with coordinates $x$ on the texture map, we assign a discrete distribution $\mathcal{P}^{(x)}$ to describe its probability of coloring with $\{c_i\}$:
\begin{align}
    p^{(x)}_{k} &= \frac{w^{(x)}_{k}}{\sum_{i=1}^{N_C} w^{(x)}_{i}}, k = 1,\dots, N_C,\label{eq:prob_map2}\\
     w^{(x)}_{i} &= \sum_{j=1}^{N_P} \exp\left(-\frac{\|x - b_{i j}\|_2}{\alpha}\right),\label{eq:prob_map1}
\end{align}
where  $p^{(x)}_{k}$ is the sampling probability of color $c_k$. According to \cref{eq:prob_map1}, the probability of a pixel $x$ colored by $c_i$ increases as it gets closer to a control point $b_{ij}$. The parameter $\alpha$ is the smoothing radius of the Voronoi diagram. When $\alpha$ approaches zero, the summation in \cref{eq:prob_map1} will be dominated by the closest control point to $x$, therefore the color of $x$ will be deterministic, which resembles the original hard version of Voronoi diagram. In practice, we define a probability map $\mathcal P$ with size $N_C\times H\times W$ for all the pixels on the texture map. We further smooth the probability map $\mathcal P$ by a uniform smoothing kernel $\mathcal S = \frac{1}{m^2}\mathbbm 1_{m\times m}$ of size $m\times m$. The smoothed probability map is then computed by a convolutional operation: $\mathcal P^\prime = \mathcal P * \mathcal S.$

\vskip 5pt
\noindent\textbf{Sampling discrete colors by Gumbel softmax.} Following the procedure stated above, we assign each pixel $x$ on the texture map to a discrete distribution $\mathcal{P}^{(x)}$ guided by a Voronoi diagram, while each pixel should be assigned with a specific color $c^{(x)}$ in the end. However, directly sampling according to $\mathcal P^{(x)}$ is not differentiable with respect to $p_i^{(x)}$. Alternatively, using softmax function directly to blend all the colors certainly can not produce discrete colors. Therefore, we leverage the Gumbel-softmax\cite{maddison2014sampling,gumbel2017} reparameterization trick to approximate the discrete sampling process.

Suppose $g_i\sim \mathrm{Gumbel}\,(0,1)$ are i.i.d. random variables drawn from the standard Gumbel distribution. Given the discrete distribution $\mathcal P^{(x)}$, we can equivalently draw the color $c^{(x)}$ by $c_k$, where
\begin{equation}
    k = \arg\max_i(g_i + \log p_{i}^{(x)}).
\end{equation}
The equivalency is guaranteed\cite{maddison2014sampling} by 
\begin{equation}
    \mathrm{Pr}(k=i) = p_{i}^{(x)}.
\end{equation}

Since the argmax operation is still non-differentiable, we instead use a softmax estimator~\cite{gumbel2017} to approximate it, such that the color $c^{(x)}$ is calculated by
\begin{equation}
    c^{(x)} = \sum_{i=1}^{N_C}{c_i \cdot \mathrm{Softmax}\left(\frac{g_i+\log p_{c_i}^{(x)}}{\tau}\right)},
    \label{eq:gumbel}
\end{equation}
where $\tau$ is the temperature coefficient. We have $\lim_{\tau \to 0}{c^{(x)}} = c_k.$
Finally, each pixel $x$ on the texture map $T$ will be colored with $c^{(x)}$. 

In order to enlarge the optimization space, we replace the random seed $g_i$ with a variable $g_i'$. Since the random seed $g_i$ can be equivalently sampled by inverse transform sampling $g = -\log(-\log(u))$ with $u\sim\mathrm{Uniform}\,(0,1)$, we define the variable $g_i$ in \cref{eq:gumbel} as
\begin{equation}
    g_i = -\log(-\log((1-\lambda) \cdot u_i^{\mathrm{(fix)}} + \lambda \cdot u_i^{\mathrm{(train)}})),
    \label{eq:gumbel_seed}
\end{equation}
where $u_i^{\mathrm{(fix)}}\sim \mathrm{Uniform}\,(0,1)$ is fixed during the whole training process, and $u_i^{\mathrm{(train)}}$ is a trainable variable clipped in range $[0, 1]$. The hyper-parameter $\lambda\in[0, 1]$ controls the ratio of the trainable variables. Putting together, we update the coordinates $\{b_{ij}\}$ and the trainable variable $\{u_i^{\mathrm{(train)}}\}$ jointly during optimization.

\subsection{Non-rigid Mesh Augmentations}

\begin{figure}[t]
    \centering
    \includegraphics[width=1.0\linewidth]{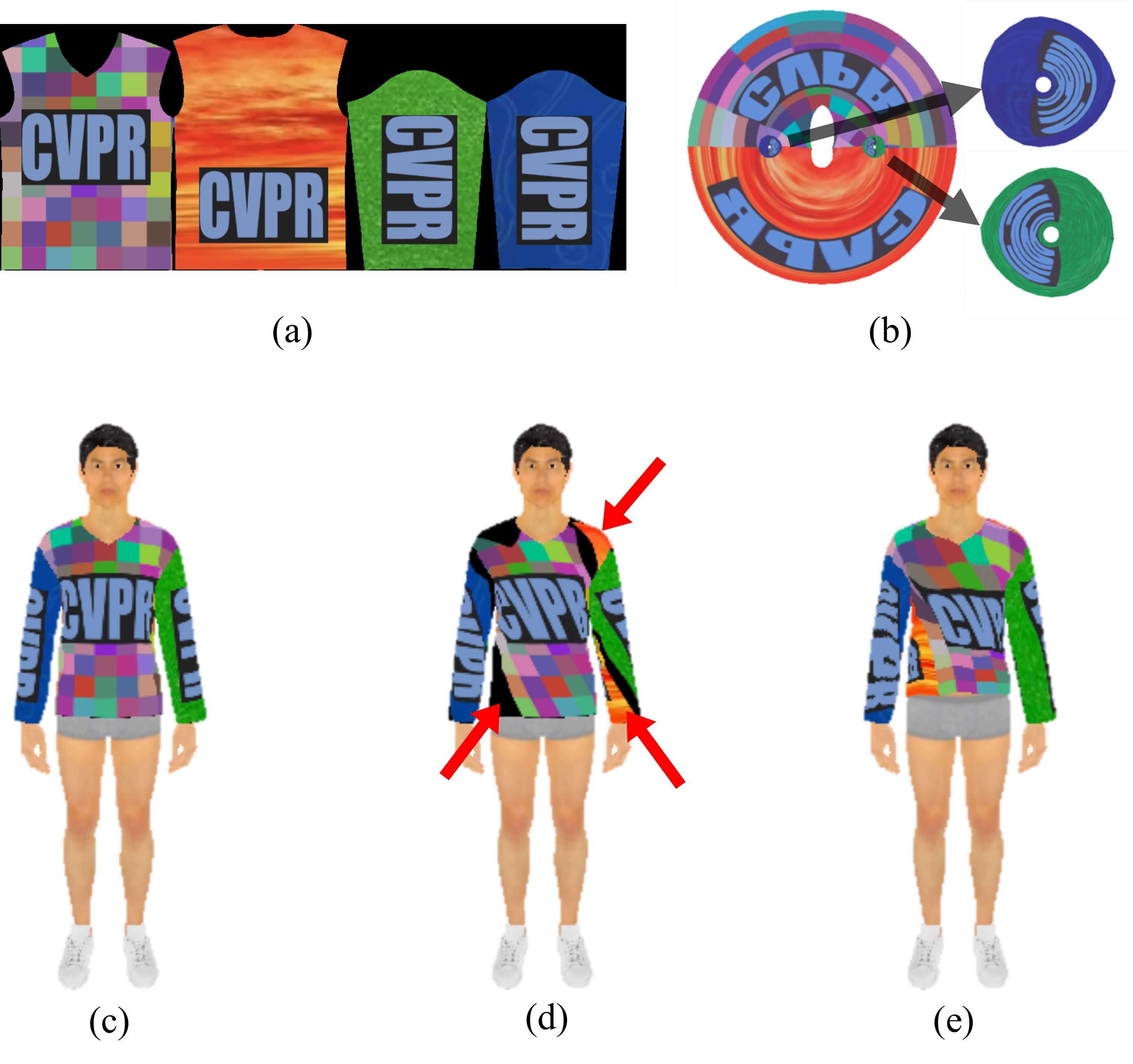}
    \caption{Visualization of the texture augmentation. (a) The texture map of a T-shirt mesh. It is geometrically plausible. (b) A texture map that is topologically plausible, where each point's neighbors in the 2D projection correspond precisely to its neighbors in the 3D mesh. (c-e) Rendered images with different warp methods on the texture map. (c) No warp. (d) Applying a mild shear strain along the texture map's vertical axis. The red arrows indicate pixels that appear at wrong places on the rendered image. (e) Our texture warp base on TopoProj.}
    \label{fig:aug_compare}
\end{figure}

According to Expectation over Transformation (EoT)~\cite{sharif2016accessorize,athalye2018synthesizing}, one can improve the robustness and the generalizability of the physical adversarial examples by applying multiple digital transformations that simulate physical transformations as augmentations during optimization. In order to efficiently simulate the physical warps and movements of the clothes, we apply two augmentations on 3D meshes before applying regular 2D augmentations on the final images. The first augmentation aims to warp the texture map of the clothes meshes based on \emph{topologically plausible projections (TopoProj)}. The second augmentation is applied on the mesh vertices' coordinates by 3D Thin Plate Spline (TPS)~\cite{tang2019augmentation}.

\vskip 5pt
\noindent\textbf{Texture warping based on topologically plausible projection.}
We first obtain the texture maps and UV coordinates of the clothes by Clo3D software\footnote{https://www.clo3d.com/}. The clothes on the 3D person models are created by pieces of flattened cloth identical to the texture maps. See (\cref{fig:aug_compare}a) for an example of a T-shirt mesh's texture map. The local distances (within triangular elements) of the mesh vertices according to 3D coordinates are thus consistent with their local distance on the 2D texture map. Therefore, we call the texture map a \emph{geometrically plausible projection (GeoProj)}. According to this local-distance-preserving property, we can produce the final clothes similar to the 3D simulated ones (\cref{fig:aug_compare}c) by printing the texture maps on fabric materials in the real world.


\begin{figure}[t]
    \center
    \includegraphics[width=1.0\linewidth]{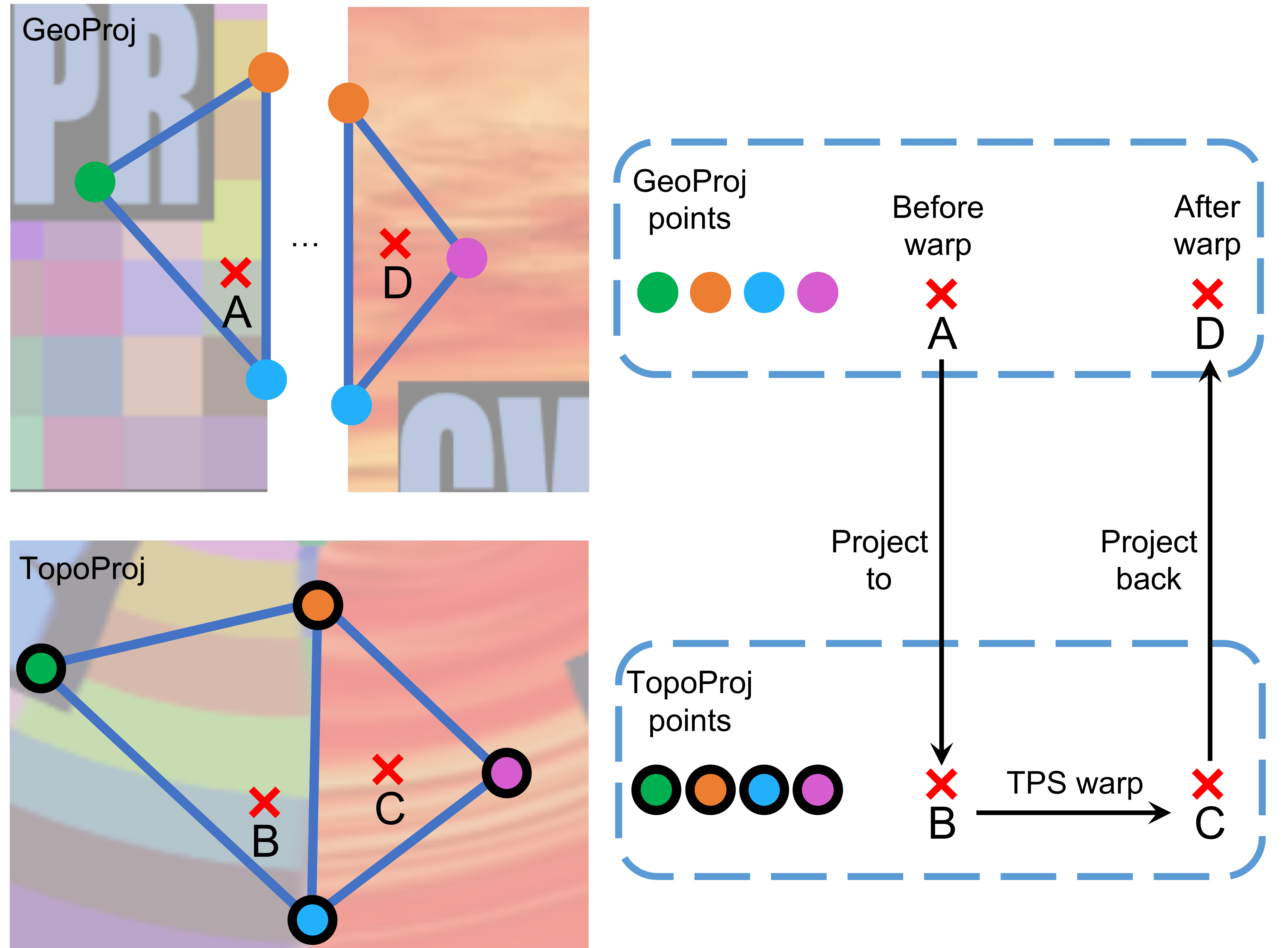}
    \caption{Illustration of the warping at the edge of two pieces. A GeoProj and a TopoProj with textures are shown on the left panel. The two pieces are far away on GeoProj but next to each other on TopoProj. Two triangle elements with blue solid lines at the eadges of the pieces are next to each other on TopoProj. The vertices on TopoProj and GeoProj with the same color are the projections of an identical vertex on the 3D mesh.}
    \label{fig:projection_detail}
\end{figure}

It is challenging to simulate the physical transformations merely by warping on the texture map, since the warping may result in transformations that are far away from the physical ones. As an example, \cref{fig:aug_compare}d shows a rendered image after applying a mild shear strain along the vertical axis in \cref{fig:aug_compare}a. Plenty of pixels on the image appear at wrong place, e.g., the black background pixels appear on the front side of the clothes, and orange backside pixels appear on the sleeves. The reason is that the coordinates of the points on the texture map cannot reflect their topological relations in the 3D mesh. For example, the bottom-left corner of the T-shirt's front side should be connected to the bottom-right corner of the T-shirt's backside, while the corresponding pixels are far away on the texture map. 

To address this problem, we propose a novel warping technique based on the TopoProj (\cref{fig:aug_compare}b), which resembles the physical transformation of the clothes (\cref{fig:aug_compare}e). A TopoProj is a projection of the mesh vertices that preserve the topological relations between vertices, which allow the pixels appear at reasonable place after the warping. See \emph{Supplementary Material} for the generation process of the TopoProj. However, we can not simply replace the GeoProj with the TopoProj, since it brings difficulties in physical realization: the local distances will no longer be consistent with that of 3D meshes, i.e., we cannot print such patterns and tailor them to produce the final clothes. Moreover, the inconsistency of the local distance will result in extremely uneven resolution of the textures.
Therefore, we leverage both GeoProj and TopoProj when applying the warping.


During the original rendering~\cite{ravi2020pytorch3d}, each pixel of the final image corresponds to a certain light path that passing through the camera. The light path may have single or multiple intersections with some triangle elements of the 3D mesh, yet we only consider the closest intersection to the camera. The barycentric coordinate of the intersection in the triangle elements thus can be calculated. Since each vertex of the triangle elements has its correspondent on the texture map, one can calculate the correspondent of the intersection point on the texture map according to its barycentric coordinate. The rendered color of the pixel thus can be calculated according to the position of the intersection point on the texture map.

In order to assign new colors to the pixels of the warped image, we apply additional projections on the coordinates of the intersection points during the rendering. \cref{fig:projection_detail} illustrates the warping pipeline. As mentioned, GeoProj and TopoProj are two projections of all the vertices in the 2D plane for a 3D mesh. 
For a point in a triangle element, we define a corespondent in each projection, whose position is determined by its barycentric coordinate. The barycentric coordinate is calculated via the original rendering, which is the same in GeoProj and TopoProj.
Specifically, we describe the warping process in five steps: (1) given an intersection point $A$ (the red cross on the left piece in \cref{fig:projection_detail}) on the GeoProj with its barycentric coordinates; (2) find its correspondent $B$ on TopoProj based on the barycentric coordinates; (3) warp the corresponding point by 2D Thin Plate Spline (TPS)~\cite{bookstein1989principal,donato2002approximate} method and get point $C$; (4) compute the new barycentric coordinates for the warped point $C$ on TopoProj (may be in a new triangle element); (5) find its correspondent $D$ on GeoProj according to the new barycentric coordinates, and compute the color of point $D$ by interpolating the texture map. The process is applied on all the pixels of the image.

The TPS warping in step (3) depends on a set of control points (See \emph{Supplementary Material} for the details). We uniformly sample the polar coordinates of each control point with a range of $[-\epsilon_r, \epsilon_r]$ and $[-\epsilon_t, \epsilon_t]$ for the radius and angle respectively.



\vskip 5pt
\noindent\textbf{Vertex augmentation by 3D TPS.} We also applied augmentation directly on the 3D vertex coordinates of the meshes by 3D TPS~\cite{tang2019augmentation}. The vertex coordinates are perturbed according to a set of control points. We uniformly perturb the control points in range $[-\epsilon_{\mathrm{TPS}}, \epsilon_{\mathrm{TPS}}]$. See \emph{Supplementary Material} for the visualization.

Together with the texture warping, we apply mesh augmentations during optimization to reduce the gap between the 3D meshes and the real-world ones. 


\vskip 5pt
\noindent\textbf{Other augmentation.} Since the colors will change when they are printed on fabric materials, we calibrate the digital color on the texture maps to the physical color following ~\cite{xu2020adversarial}. See \emph{Supplementary Material} for the details. During 3D rendering, we sample the viewing angles of the camera adaptively according to the mean confidence score of the target bounding boxes, where the angles with higher scores are more likely to be sampled. We also choose the simulated lights from ambient lights, directional lights, and point lights uniformly at random. Moreover, we apply other image augmentations on the rendered images following previous works~\cite{thys2019fooling,wang2020can}, such as randomizing the scales, positions, contrast and brightness.


\subsection{Adversarial Loss Function}
\label{subsection:loss}

In this section, we present the objective functions for attacking detectors.
\vskip 5pt
\noindent\textbf{Detection loss.} Object detectors predict bounding boxes with confidence scores. Since our goal is to evade the detectors from detecting humans, we minimize the confidence score of the person class in the box which has the maximum Intersection over Union (IoU) score with the ground truth. For each input $x$, suppose that the victim detector $\mathcal D$ outputs a set of bounding boxes $b_i^{(x)}$, each with a confidence $\mathrm{Conf}_{i}^{(x)}$. We define the detection loss as
\begin{equation}
    \mathcal L_{\textit{det}} = \sum_x{\mathrm{Conf}_{i^*}^{(x)}},
    i^* = \arg\max_i{\mathrm{IoU}(\mathrm{gt}^{(x)}, b_i^{(x)})},
\end{equation}
where $\mathrm{gt}^{(x)}$ stands for the ground truth bounding box of the foreground person on the input image $x$.

\vskip 5pt
\noindent\textbf{Concentration loss for camouflage texture.} 
In order to increase the stability of the camouflage texture generation, we prevent the polygons from being too small by introducing a concentration loss that encourages control points to move away from each other:
\begin{equation}
    \mathcal L_{con} = \sum_{j=1}^{N_C} \sum_{1 \leq k_1 < k_2 \leq N_P} \exp\left(-\frac{\|b_{jk_1} - b_{jk_2}\|^2}{\sigma^2}\right),
\end{equation}
where $\sigma$ is a constant.


The total adversarial loss for minimization is
\begin{equation}
    \mathcal L = \mathcal L_{det} + \alpha_{con}\mathcal L_{con},
\end{equation}
where $\alpha_{con}$ is the weight between the two losses.


\begin{table*}[t]
\centering
\small
\begin{tabular}{ccccccccc}
\toprule
Images               \
& \begin{subfigure}[t]{0.08\textwidth}\centering\includegraphics[width=\textwidth]{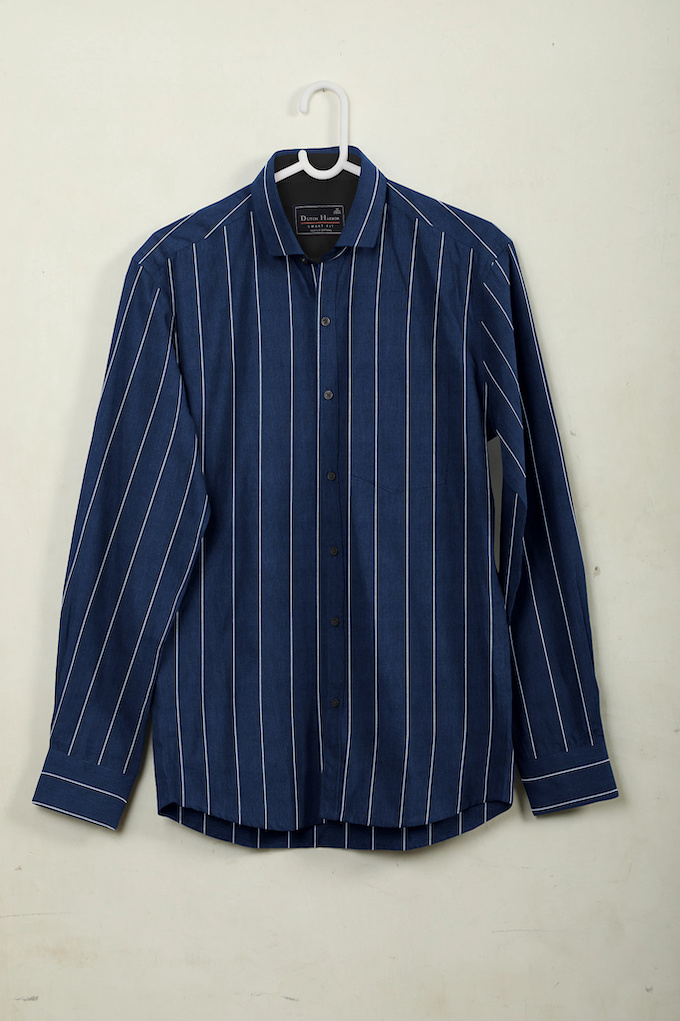}\end{subfigure} \
& \begin{subfigure}[t]{0.08\textwidth}\centering\includegraphics[width=\textwidth]{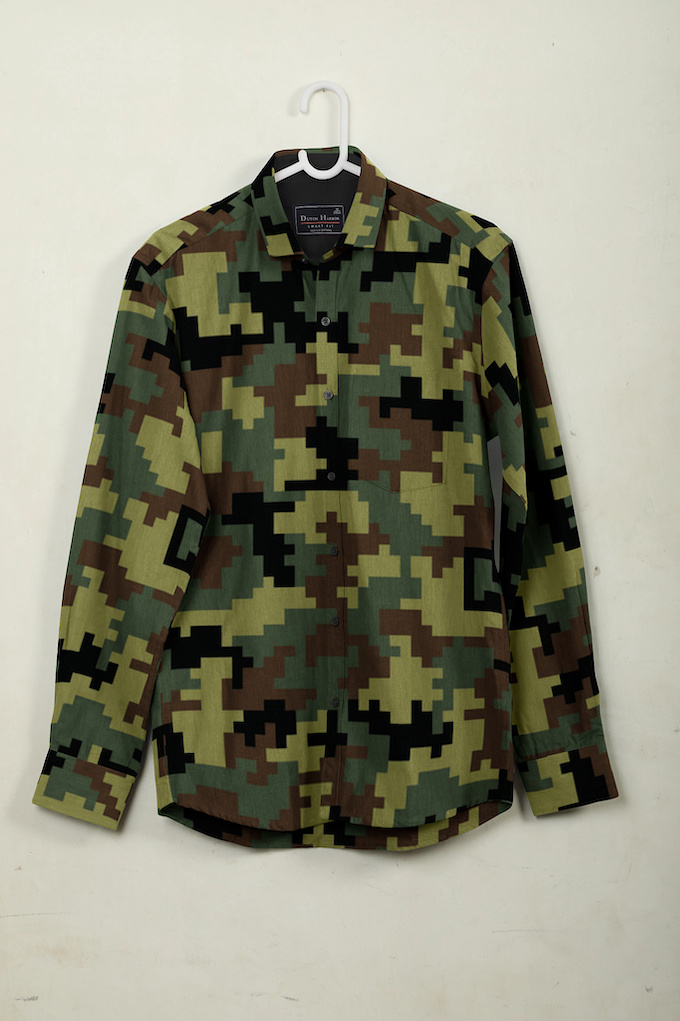}\end{subfigure} \
& \begin{subfigure}[t]{0.08\textwidth}\centering\includegraphics[width=\textwidth]{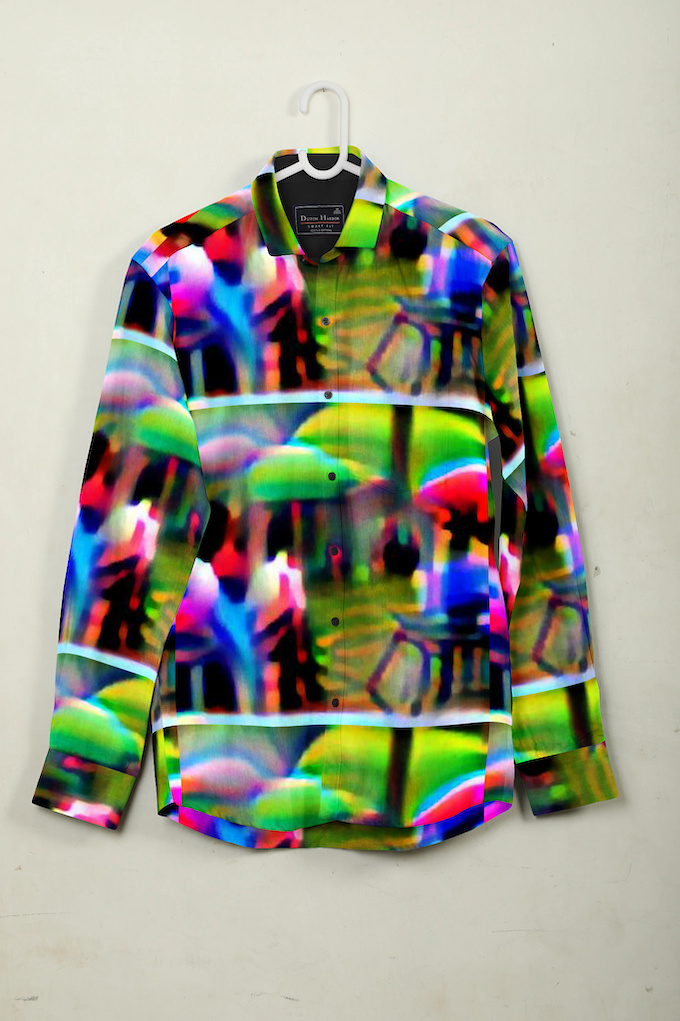}\end{subfigure} \
& \begin{subfigure}[t]{0.08\textwidth}\centering\includegraphics[width=\textwidth]{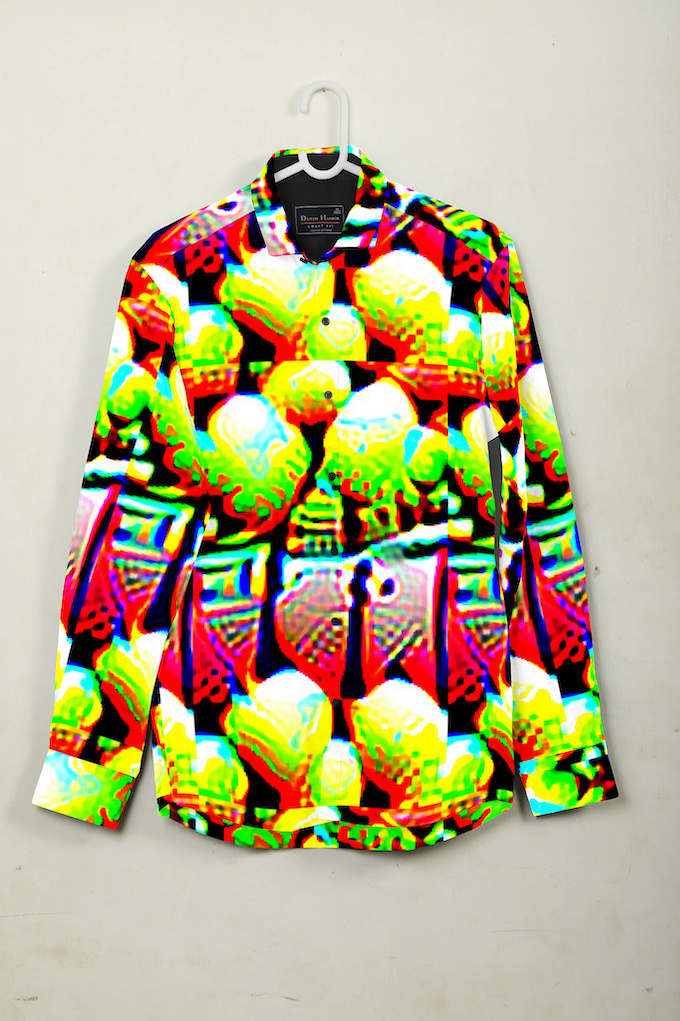}\end{subfigure} \
& \begin{subfigure}[t]{0.08\textwidth}\centering\includegraphics[width=\textwidth]{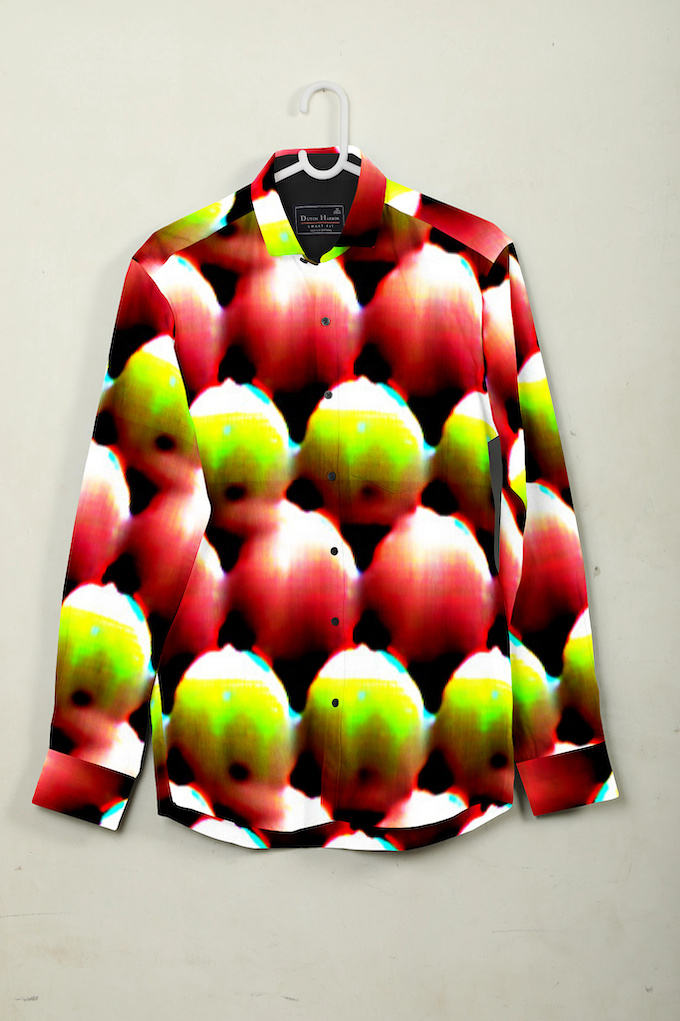}\end{subfigure} \
& \begin{subfigure}[t]{0.08\textwidth}\centering\includegraphics[width=\textwidth]{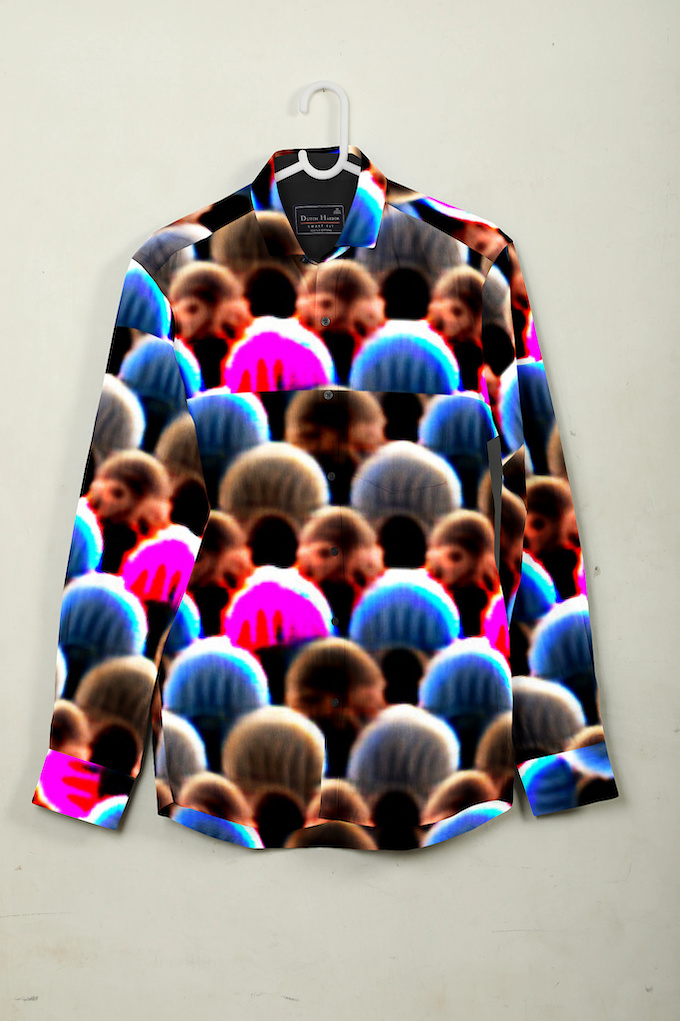}\end{subfigure} \
& \begin{subfigure}[t]{0.08\textwidth}\centering\includegraphics[width=\textwidth]{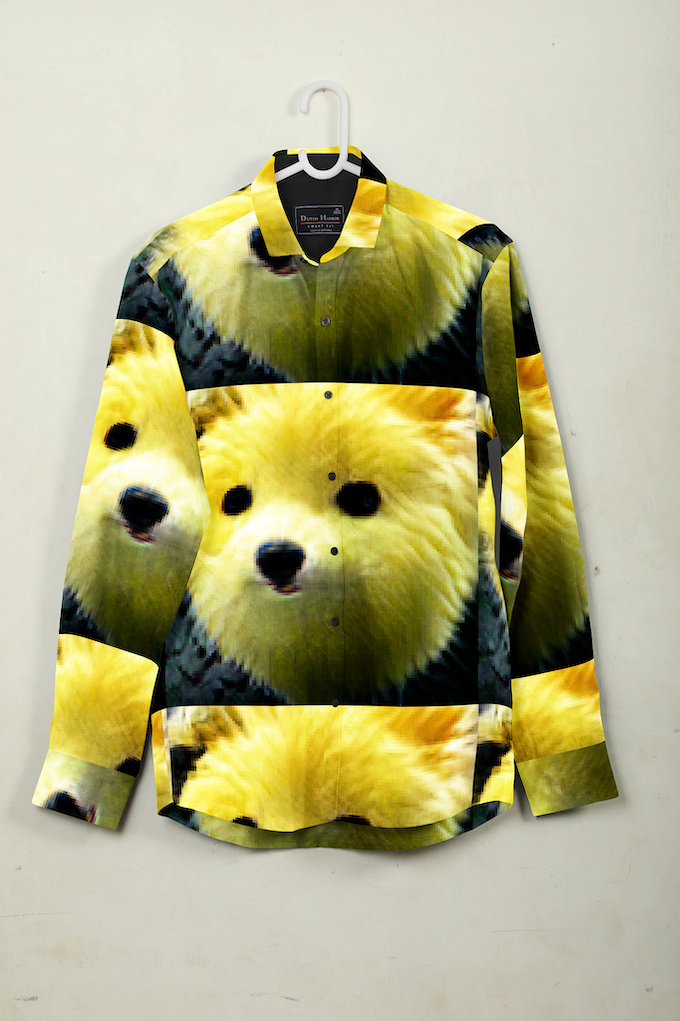}\end{subfigure} \
& \begin{subfigure}[t]{0.08\textwidth}\centering\includegraphics[width=\textwidth]{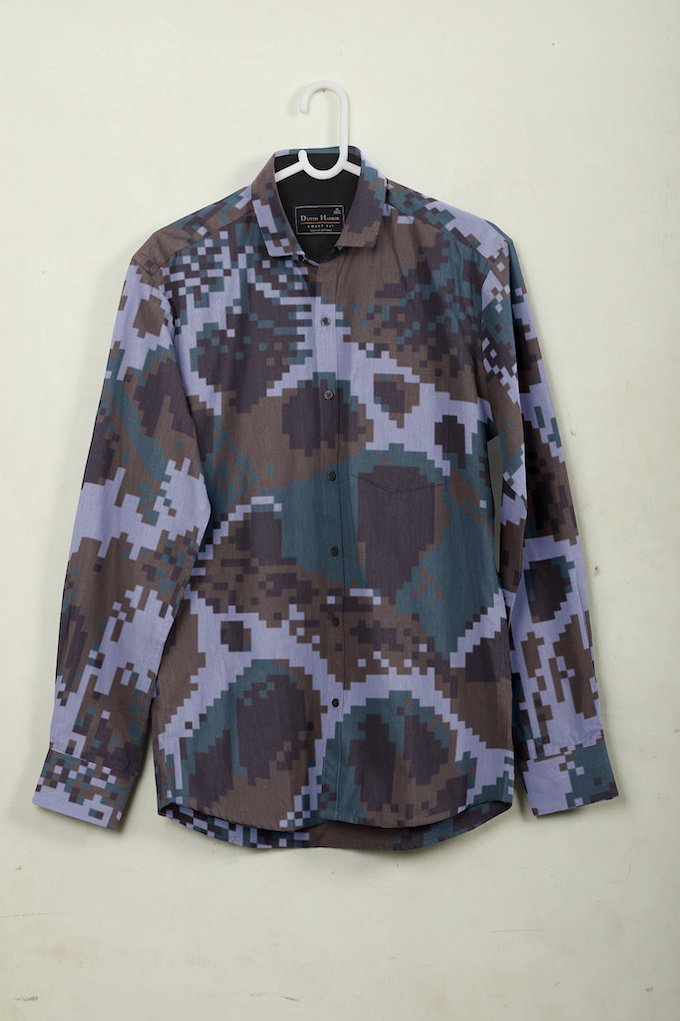}\end{subfigure} \\
\midrule
Score & $6.08\pm1.00$ & $5.05\pm1.39$ & $2.05\pm1.06$ & $1.75\pm1.09$ & $1.72\pm0.98$ & $1.69\pm0.95$ & $2.54\pm1.23$ & $4.89\pm1.39$ \\
\midrule
Source               & \makecell[c]{common\\texture}  & \makecell[c]{common\\camouflage}  & \makecell[c]{AdvPatch\\ \cite{thys2019fooling}} & \makecell[c]{ AdvTshirt\\ \cite{xu2020adversarial}} & \makecell[c]{AdvTexture\\yolo\cite{hu2022adversarial}}  & \makecell[c]{AdvTexture\\faster\cite{hu2022adversarial}} & \makecell[c]{NatPatch\\ \cite{hu2021naturalistic}} & \makecell[c]{AdvCaT\\(ours)} \\
\bottomrule
\end{tabular}
\caption{Subjective test using a 7-level Likert scale (1 = not natural at all to 7 = very natural).}
\label{tab:human_test}
\end{table*}

\section{Experiments}
\subsection{Experimental Setup}
\vskip 5pt
\noindent\textbf{Subjects.}
Three actors (age mean: $26.3$; age range: $25-27$; height range: $175-178\;\mathrm{cm}$) are recruited to collect physical test data. We also recruited $93$ subjects (age mean: $30.2$; age range: $18-57$) to evaluate the naturalness score of different clothes. The recruitment and study procedures were approved by the Department of Psychology Ethics Committee, Tsinghua University, Beijing, China.

\vskip 5pt
\noindent\textbf{Baseline methods}
According to the previous work~\cite{hu2022adversarial}, the Attack Success Rates (ASRs) of the adversarial clothes printed with isolated patches will drop catastrophically when the viewing angle changes. Printing repetitively tiled patches on the clothes is helpful to prevent the ASRs from dropping. Therefore, we tiled the patches produced by patch-based attacks for fair comparison. We mainly evaluated three patch-based attacks \emph{AdvPatch}\cite{thys2019fooling}, \emph{AdvTshirt}\cite{xu2020adversarial}, \emph{NatPatch}\cite{hu2021naturalistic}, and a texture-based attack, \emph{AdvTexture}\cite{hu2022adversarial}. We also evaluate \emph{RandColor}, a random texture with random colors in a lattice, and \emph{RandCaT}, a random camouflage texture pattern.


See \emph{Supplementary Material} for the datasets, target detectors, evaluation metric and the implementation details.

\subsection{Naturalness Score by Subjective Evaluation}

Following Hu et al.~\cite{hu2021naturalistic}, we conducted a subjective evaluation on the naturalness score of the adversarial clothes. For a fair comparison, we applied different patterns on an identical garment model using FAB3D\footnote{https://tri3d.in/}. We showed eight pictures of different T-shirts  (\cref{tab:human_test}) aggregated on a scrollable page in random orders to the subjects and required them to give a naturalness score for each picture using a 7-level Likert scale (1 = not natural at all to 7 = very natural).

As shown in \cref{tab:human_test}, the naturalness score of AdvCat targeting Faster RCNN ($4.89$) is significantly higher than those of the other five adversarial patterns ($p<0.001$, student's t-test), and is close to the scores of the control group with common textures (the second column, $6.08$ and the third column, $5.05$ in the table). 

\subsection{Digital World Results}
\label{sec:digital}

\begin{table}[t]
\centering
\small
\begin{tabular}{lllll}
\toprule
Method            & $\mathrm{IoU}0.01$  & $\mathrm{IoU}0.1$ & $\mathrm{IoU}0.3$ & $\mathrm{IoU}0.5$  \\
\midrule
RandColor      & 0.13    & 0.13  & 0.13    & 0.17    \\
RandCaT      & 1.02    & 1.02 & 1.04    & 1.10    \\
AdvPatch     & 69.33    & 72.27   & 75.80    & 85.97    \\
NatPatch     & 42.47    & 43.66   & 45.41    & 67.40    \\
AdvTexture     & 1.44    & 21.73   & 87.05    & \textbf{99.98}    \\
\midrule
AdvCaT (ours)      & \textbf{95.18}   & \textbf{99.21} & \textbf{99.40}    & 99.52     \\ 
\bottomrule
\end{tabular}
\caption{ASRs/\% of different methods targeting Faster RCNN in the digital world.}
\label{tab:digital_all}
\end{table}

\vskip 5pt
\noindent\textbf{Evaluation with different IoU threshold.}
We noticed that the IoU threshold $\tau_{\mathrm{IoU}}$ during evaluation is usually set to $0.5$ according to previous works~\cite{thys2019fooling,hu2022adversarial} since they mainly evaluate their adversarial patches or textures on the datasets that contains multiple people, e.g. Inria dataset~\cite{dalal2005histograms}. On such datasets, a relatively high threshold can prevent confusing the boxes of overlapping objects. However, the high threshold could result in an overestimation of the attack's effectiveness. The target detector may output a considerably large bounding box with an IoU score smaller than the threshold, which still provides strong evidence of having detected the person. See \emph{Supplementary Material} for examples. Therefore, we evaluated the ASRs with different IoU thresholds $0.01, 0.1, 0.3$, and $0.5$. See \cref{tab:digital_all} for the ASRs of different methods targeting Faster RCNN. According to \cref{tab:digital_all}, the ASR of the AdvTexture is slightly higher than our method with an IoU threshold of $0.5$, while it decreases significantly when the IoU threshold decreases. On the contrary, the ASRs of our method, AdvCaT is consistently high with different IoU threshold, even when the threshold equals to $0.01$, which indicates the strong adversarial effectiveness of AdvCaT. Since an IoU threshold of $0.01$ is too small that may introduce undesirable noises, we report the ASRs with IoU threshold $0.1$ in the rest of our paper unless explicitly mentioned.

\begin{figure}[t]
    \centering
    \includegraphics[width=1.0\linewidth]{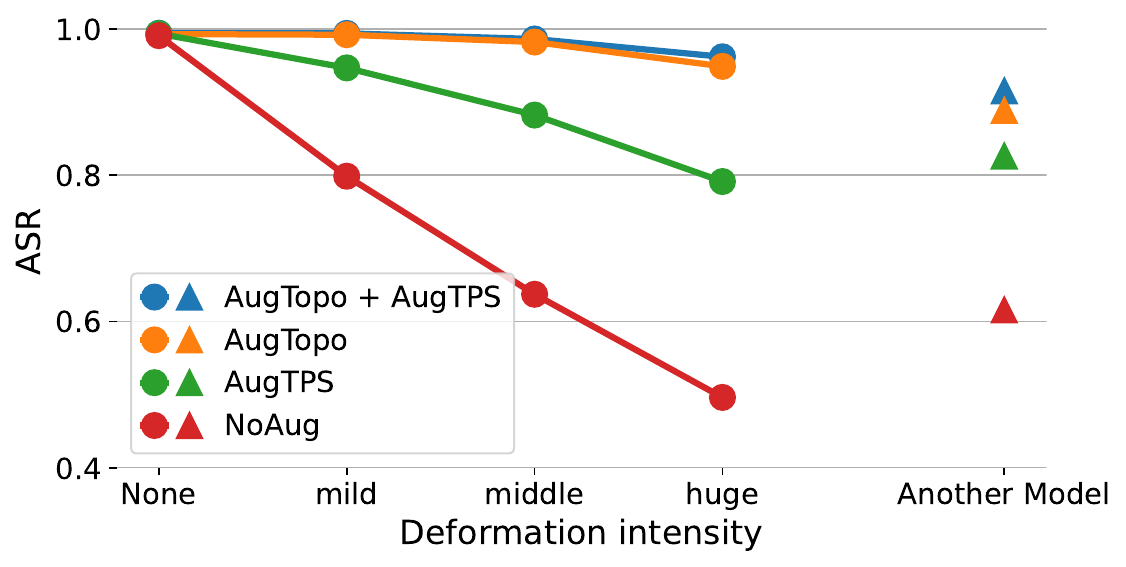}
    \caption{ASR targeting Faster RCNN versus deformation intensity for different augmentation strategies during training. \emph{Another Model} denotes ASRs of the patterns when they are applied on an unseen model without any 3D deformation.}
    \label{fig:aug_plot}
\end{figure}

\vskip 5pt
\noindent\textbf{Ablation Study of Augmentation Strategies} In order to investigate the effect of different 3D model augmentation strategies on the generalizability of the optimized patterns, we optimized four AdvCaT patterns targeting Faster RCNN with different 3D model augmentation strategies
The first pattern used no augmentation on 3D meshes, denoted by \emph{NoAug}. The second pattern used 3D TPS augmentation, denoted by \emph{AugTPS}. The third pattern used topologically plausible projection, denoted by \emph{AugTopo}. The final one, \emph{AugTPS+AugTopo}, incorporated both augmentations. Moreover, we used four different intensity of 3D mesh deformation during evaluation, which are denoted by \emph{None} (no deformation), \emph{Mild} ($(\epsilon_r, \epsilon_t, \epsilon_{\mathrm{TPS}} )=(0.1, 50, 0.15)$), \emph{Middle} ($(\epsilon_r, \epsilon_t, \epsilon_{\mathrm{TPS}} )=(0.1, 65, 0.22)$), and \emph{Huge}$(\epsilon_r, \epsilon_t, \epsilon_{\mathrm{TPS}} )=(0.1, 80, 0.3)$, respectively. Note the hyper-parameter used during training was the same as \emph{Mild}. See \cref{fig:aug_plot} for the ASRs of the patterns with different augmentation strategies and deformation intensities. The ASRs of the patterns applied on a new 3D person without any 3D deformations are also plotted in the figure.

As shown in \cref{fig:aug_plot}, the ASR of \emph{NoAug} drops significantly when the deformation intensity increases, which implies its catastrophic drop of the adversarial effectiveness in the real world. Using 3D TPS alone can be better, but it still suffers from a considerable drop under huge deformation intensity. The ASRs of \emph{AugTopo} that only use TopoProj are high even when the deformation intensity is huge. Combining 3D TPS with TopoProj is slightly better than only using TopoProj. The ASRs of different strategies evaluated on a new unseen 3D model are consistent with the previous observation, which indicates the good generalization ability of the pattern using both augmentations.


\begin{table}[t]
    \centering
    \small
    \begin{tabular}{lccc}
    \toprule
         Detectors &  Faster RCNN & YOLOv3 & DDETR\\
    \midrule
        Random & 0.85 & 3.31 & 5.76\\
        AdvCaT & 99.36 & 94.53 & 88.77\\
    \bottomrule
    \end{tabular}
    \caption{ASRs/\% targeting different detectors in the digital world.}
    \label{tab:different_detector}
\end{table}

\vskip 5pt
\noindent\textbf{Attacking different detectors.} We optimized camouflage patterns to attack different detectors including YOLOv3~\cite{redmon2018yolov3}, FasterRCNN~\cite{ren2016faster} and deformable DETR~\cite{Zhu2021Deformable} and show their ASRs in \cref{tab:different_detector}. We also used the trained patterns to attack other detectors to study their transferability. The ASR of the AdvCaT trained to attack Faster RCNN was relatively high when targeting MaskRCNN ($92.22\%$) and Deformable DETR ($65.11\%$), but relatively low when targeting YOLOv3 ($23.26\%$). See \emph{Supplementary Material} for the visualization of these AdvCaTs and the full transfer study.
\vskip 5pt
\noindent\textbf{Parameter sensitivity.} We varied the value of $\lambda$ in \cref{eq:gumbel_seed} during optimization. When $\lambda$ increased, The ASR increased, while the AdvCaT became less like a camouflage pattern. In addition, we optimized AdvCaT with different styles by using various color combinations ${c_i}$, all of which achieves high ASRs targeting Faster RCNN. See \emph{Supplementary Material} for details of these experiments.

\subsection{Physical World Results}



\begin{figure}[t]
\centering
\includegraphics[width=1.0\linewidth]{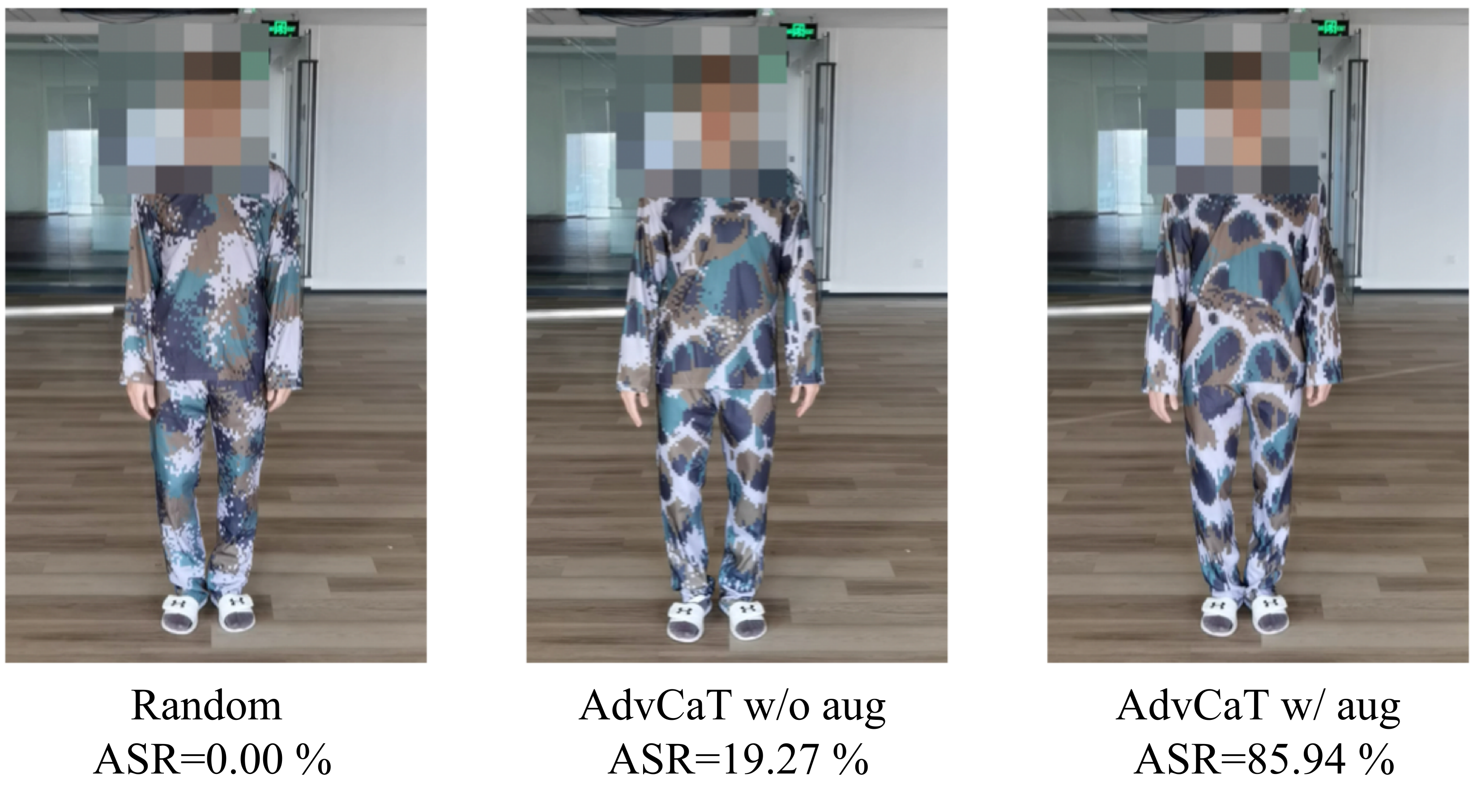}
\caption{Adversarial clothes covered with different patterns in the physical world.}
\label{fig:physical_eval}
\end{figure}

\begin{figure}[t]
\centering
\includegraphics[width=1.0\linewidth]{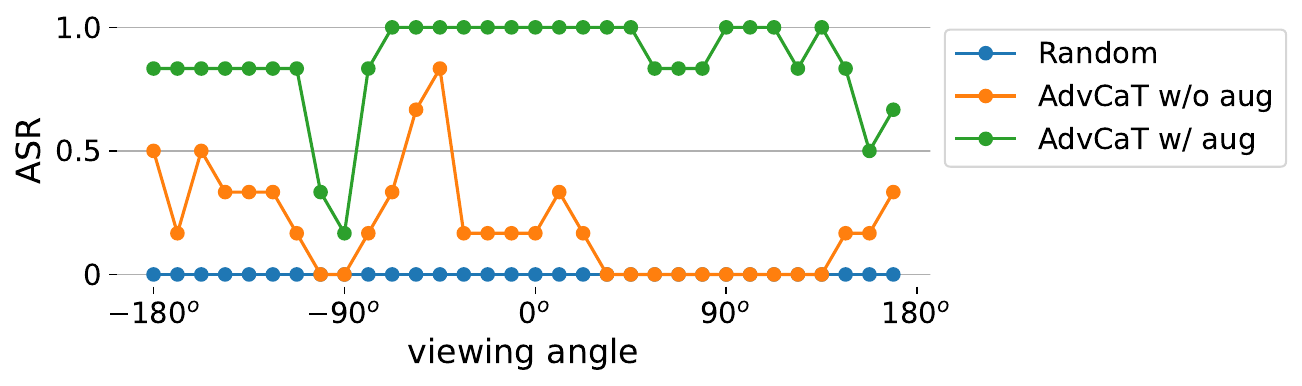}
\caption{ASRs at different viewing angles in the physical world. The actors were facing the camera when the viewing angle equals to $0^\circ$.}
\label{fig:angle_physical}
\end{figure}

We produced three clothes covered with different AdvCaT patterns in the physical world. We cropped the different parts of the clothes from the texture map and printed them on fabric materials. These pieces were then tailored to wearable adversarial clothes. In \cref{fig:physical_eval} we visualized the clothes and presented their ASRs targeting Faster RCNN, where \emph{Random} denotes the clothes covered with random camouflage textures; \emph{AdvCaT w/o aug} and \emph{AdvCaT w aug} denote the clothes that covered with AdvCaT with or without mesh augmentation (i.e. TopoProj and 3D TPS) during optimization, respectively. The ASR of \emph{AdvCaT w aug} ($85.94\%$) was significantly higher than those of \emph{AdvCaT w/o aug} ($19.27\%$) and \emph{Random} ($0.00\%$). 

\cref{fig:angle_physical} shows the ASRs at different viewing angles, indicating strong attack ability of the AdvCaT clothes. In addition, we found that our designed clothes were relatively robust to the environment change. When the distance between the actor and the camera was less than $4\;\mathrm{m}$, the ASR stayed high (above $61.5\%$). See \emph{Supplementary Material} for details of these experiments. We also provide a video demo in \emph{Supplementary Video}.

\section{Conclusion}
We proposed to optimize clothes textures via 3D modeling to produce natural-looking adversarial clothes that are adversarially effective at multiple viewing angles. 
The adversarial T-shirt with AdvCaT patterns has a high naturalness score in a subjective test evaluated by a group of subjects. Experimental results indicate that our adversarial clothes can hide people from detectors at multiple viewing angles with high ASRs in the digital and physical world.

\vskip 5pt
\noindent\textbf{Limitations}
Though the AdvCaT patterns sometimes have a relatively high ASR targeting unseen detectors, their transferability is not universal, since the ASRs targeting some certain detectors are not very good. One can use model ensemble to improve their transferability. 

\section*{Acknowledgement}
This work was supported by the National Natural Science Foundation of China (Nos. U19B2034, 62061136001, 61836014).

{\small
\bibliographystyle{ieee_fullname}
\bibliography{myegbib}
}

\renewcommand\thefigure{S\arabic{figure}}
\renewcommand\thetable{S\arabic{table}}
\setcounter{figure}{0}
\setcounter{table}{0}
\onecolumn

\appendix

\section{Supplementary Methods}
\subsection{Topologically Plausible Projection Generation}

\begin{figure}[h]
    \centering
    \includegraphics[width=0.6\linewidth]{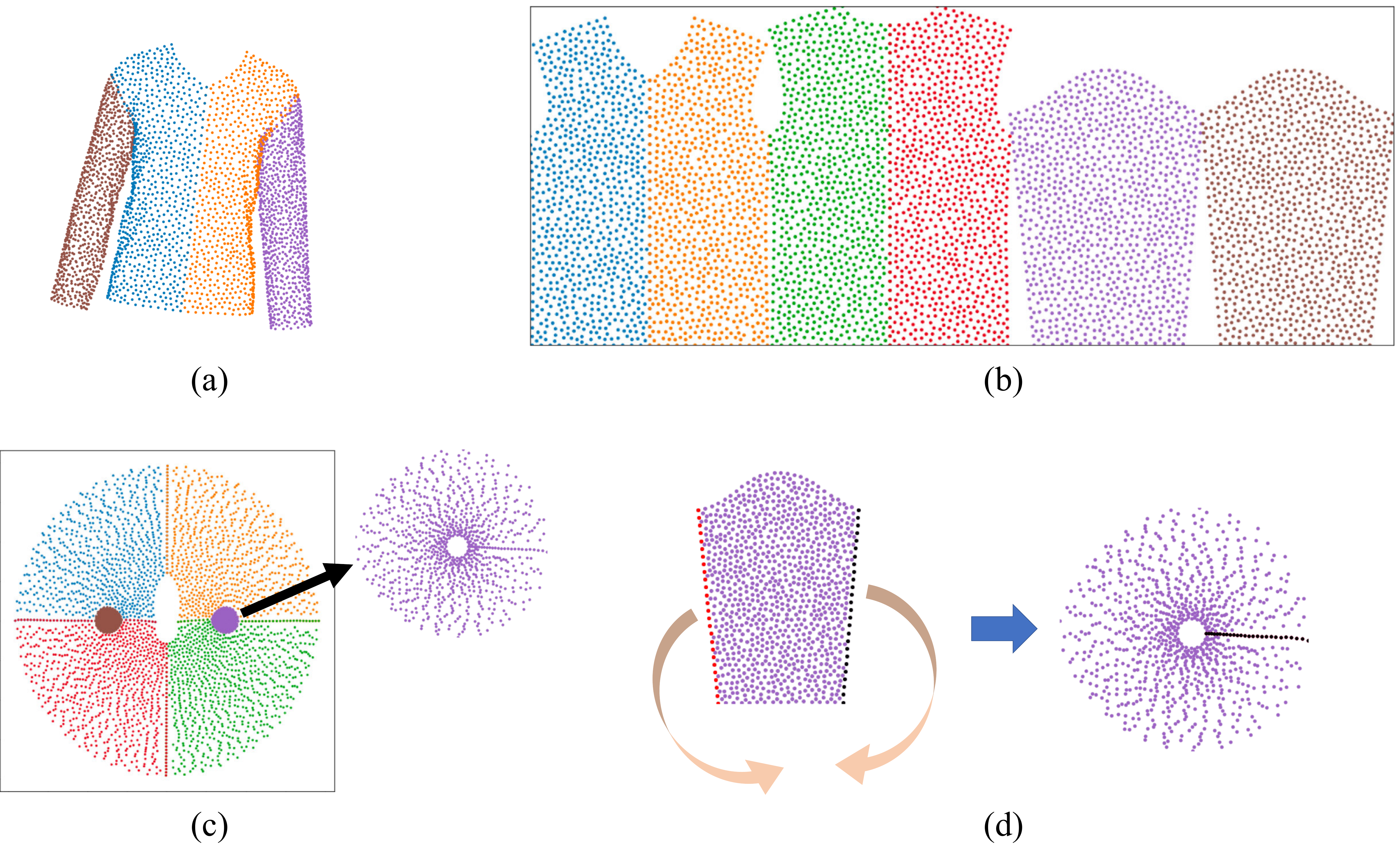}
    \caption{Visualization of different projections. (a) The 3D vertex points of a T-shirt mesh at the front view. (b) A GeoProj. (c) A TopoProj. (d) A TopoProj of a sleeve can be created from its GeoProj by Zipping. Each pair of the points on GeoProj that supposed to be identical are colored red and black, respectively. The points on the left (red) and right (black) boundaries in the left panel are identical according to 3D mesh and therefore have the same coordinates in TopoProj (the right panel). 
    }
    \label{fig:verts_plot}
\end{figure}

See \cref{fig:verts_plot}a for the visualization of the vertices of a T-shirt mesh, and \cref{fig:verts_plot}b-c for the visualization of a GeoProj and a TopoProj. We generate the topological plausible projection (TopoProj) by a process named \emph{Zipping}. When 3D vertex points are mapped to TopoProj, the connectivity of all point pairs remains unchanged. However, as for GeoProj, only the connectivity of the inner points in each pieces is unchanged, while some separated boundary point pairs are connected according to the 3D mesh. Therefore, we use Zipping to generate TopoProj by connecting the point pairs in GeoProj. See \cref{fig:verts_plot}d for an example.

\begin{figure}[t]
    \centering
    \includegraphics[width=0.5\linewidth]{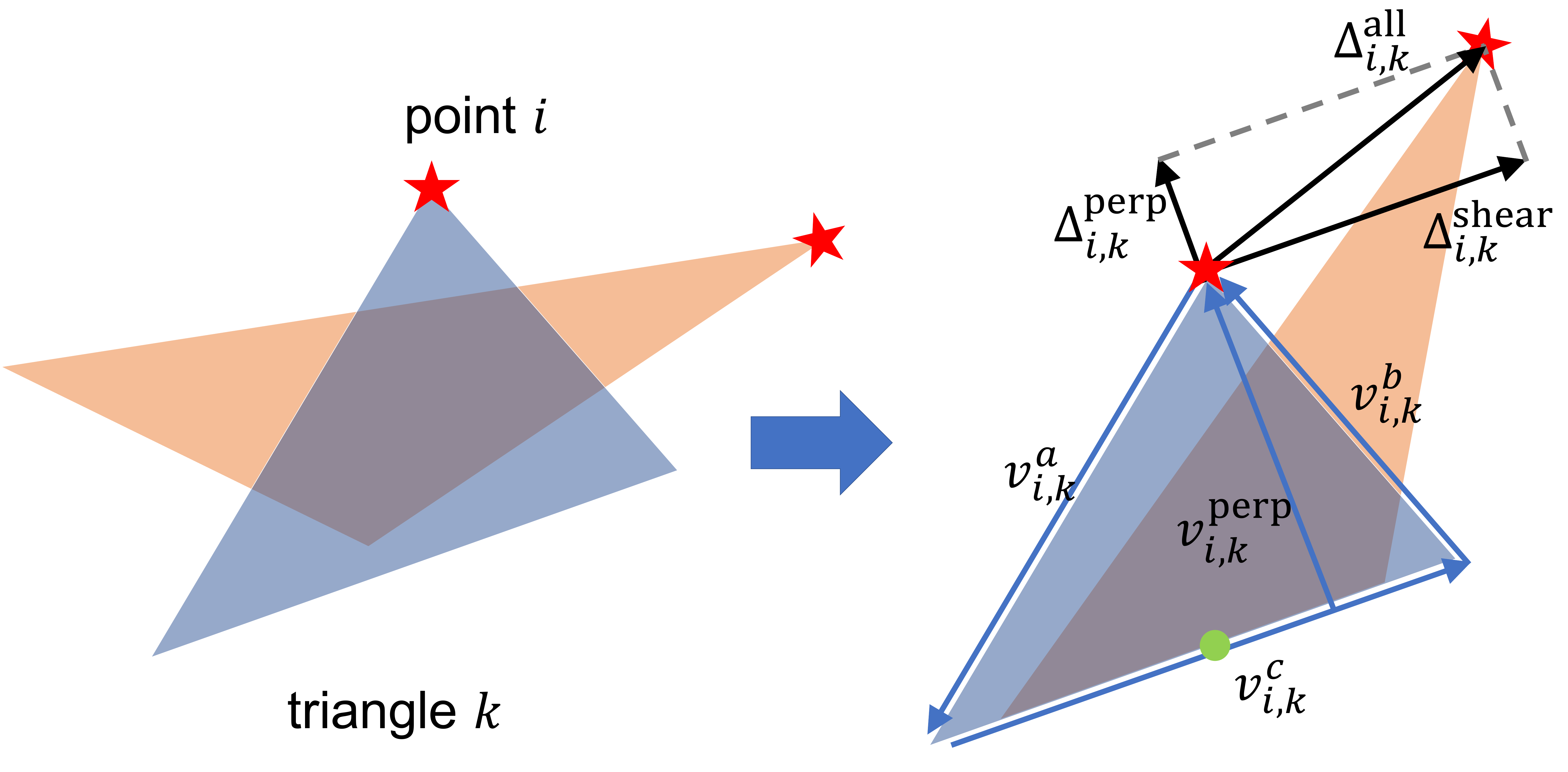}
    \caption{Illustration of the forces. The light Salmon orange triangle and the blue triangle denotes the triangle element before and after deformation, respectively. For a specific point, denoted by the red star, we first rotate and translate the triangles to align its bottom edges in the two triangles with overlapped midpoints (colored green). The force is then defined proportional to the displacement vector $\Delta_{i,k}^{\mathrm{all}}$ starting from the point after deformation and ending at the point before deformation.}
    \label{fig:force}
\end{figure}

Zipping is a continuous transformation from GeoProj to TopoProj inspired by physical mechanisms of elastic stress. Intuitively, Zipping is to pull the point pairs (red and black points in \cref{fig:verts_plot}d) slowly, until they are overlapped. The triangle elements will be deformed as even as possible, while their chiralities are not supposed to be flipped during zipping. Specifically, we update the coordinates of the points based on carefully designed forces. This force is an inner force caused by the triangle (like a spring) that intended to push the point back to \emph{restore} the triangle's original shape. It restricts the shape of the entire frame from changing too much during Zipping. As illustrated in \cref{fig:force}, the force component $F_{i,k}$ on point $i$ by triangle element $k$ is proportional to the relative displacement $\Delta_{i,k}^{\mathrm{all}}$ of this point in the triangle element. Since each point belongs to multiple triangle elements, the resultant force is
\begin{align}
  \label{eq:force}
  F_{i} &= \sum_k{F_{i,k}} = \sum_k{\frac{\Delta_{i,k}^{\mathrm{all}}}{\left\|v_{i,k}^\mathrm{perp}\right\|_2} },\\
  \label{eq:altitude}
  v_{i,k}^\mathrm{perp} &= \frac{(v_{i,k}^b, v_{i,k}^c)v_{i,k}^a - (v_{i,k}^a, v_{i,k}^c)v_{i,k}^b}{(v_{i,k}^c, v_{i,k}^c)},
\end{align}
where $(\cdot, \cdot)$ is the inner production and $v_{i,k}^a$, $v_{i,k}^b$ and $v_{i,k}^c$ are the edge vectors of the triangle element illustrated in \cref{fig:force}. $v_{i,k}^\mathrm{perp}$ is the altitude from the edge $v_{i,k}^c$ to the point $i$, also illustrated in \cref{fig:force}. We divide the forces by the norm of the altitude $\left\|v_{i,k}^\mathrm{perp}\right\|_2$ in \cref{eq:force} to prevent the triangle element from flipping.
Moreover, the points in each point pair (colored red and black in \cref{fig:verts_plot}d) are subjected to attractive forces that point to each other. We then iteratively update the coordinates of each point according to the resultant force. Suppose that the resultant force of the point $i$ at step $t$ is $F_i^{(t)}$, and its current coordinate is $x_i^{(t)}$. We update the coordinate by
\begin{equation}
  x_i^{(t + 1)} = x_i^{(t)} + \beta^{(t)} * F_i^{(t)},
\end{equation}
where $\beta^{(t)}$ is an adaptive time interval that prevent the chirality of the triangles from being flipped. Specifically, we have
\begin{align}
  \beta^{\mathrm{(t)}} &= \gamma\min{\left(\mathcal{S}^{(t)}_\beta \cup \{\beta_{\mathrm{max}}\}\right)}, \\
  \mathcal{S}_\beta ^{(t)} &={\left\{\beta\mid\beta>0,\text{ and } \exists i,k, \text{ s.t. } {v_{i,k}^\mathrm{perp} = 0}\text{ when coordinates } x_i = x_i^{(t)} + \beta * F_i^{(t)}\right\}},
\end{align}
where we used $\gamma=0.5$ and $\beta_{\mathrm{max}}=0.1$ in practice.

\begin{figure}[t]
    \centering
    \includegraphics[width=0.9\linewidth]{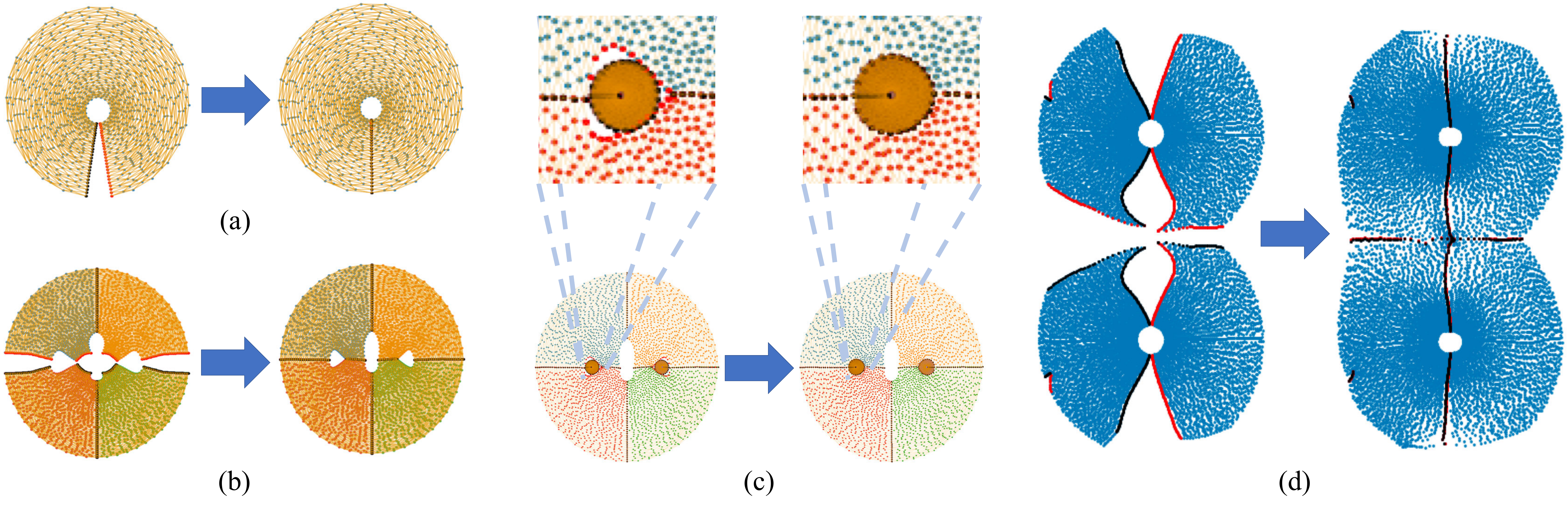}
    \caption{Visualization of Zipping when generating different parts of the clothes. The left panels of all the sub-figures are the initial states of the points, and the right panels are the points after Zipping.  (a) One of the sleeves. (b) Zipping the front side of the T-shirt with the back side. (c) Zipping the sleeves with the front side and back side of the T-shirt. (d) The trousers.}
    \label{fig:projection_vis}
\end{figure}

With a proper initialization of the points' coordinates (by bending the pieces from GeoProj), we generated the TopoProj for T-shirt and trouser meshes. See the visualization in \cref{fig:projection_vis}.

\subsection{TPS Warping on TopoProj}

\begin{figure}[t]
    \centering
    \includegraphics[width=0.8\linewidth]{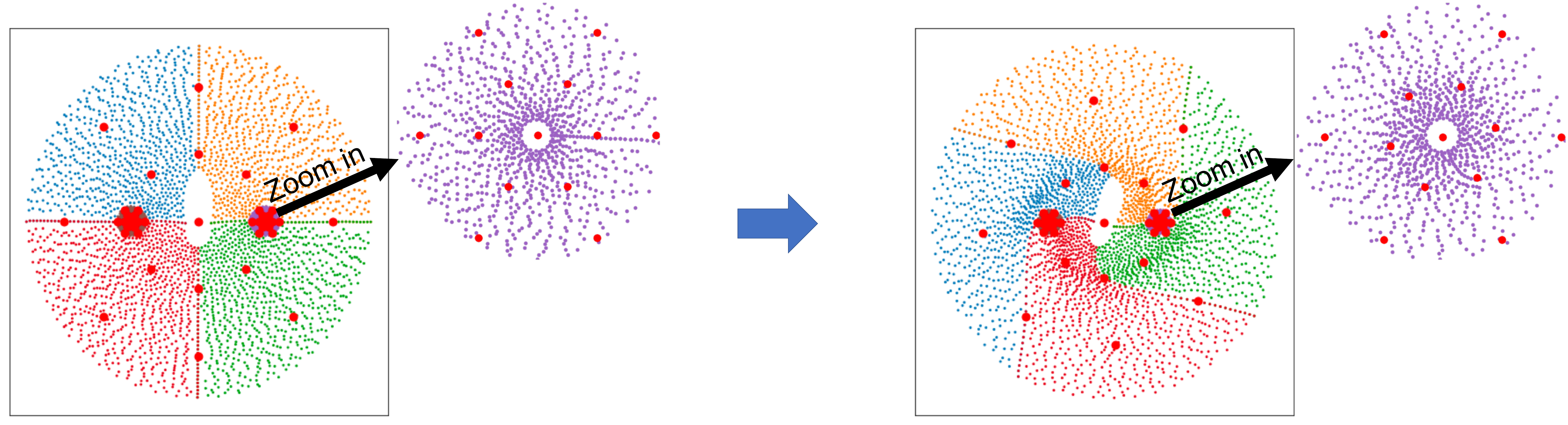}
    \caption{The warping on TopoProj of the T-shirt mesh by 2D TPS. Left: the TopoProj before warping; Right: the TopoProj after warping. The large red dots denote the control points for 2D TPS warping. We zoom in the locations around the sleeves to see the details}
    \label{fig:projection}
\end{figure}

We initialize a set of control points and uniformly perturb the polar coordinates of each control point. The warped coordinates of all the other points are calculated according to the control points, as shown in \cref{fig:projection}.

\subsection{Visualization of 3D TPS}
\begin{figure}[t]
    \centering
    \includegraphics[width=0.5\linewidth]{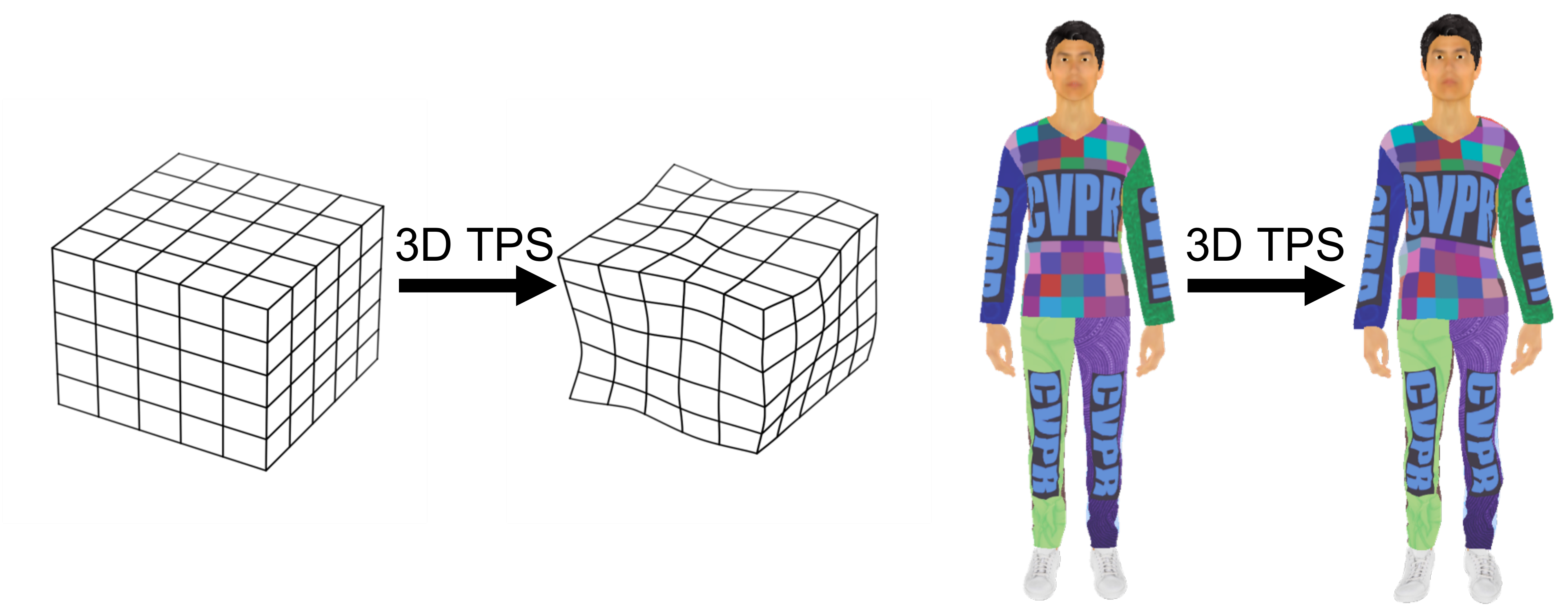}
    \caption{Visualization of 3D TPS warping. We apply 3D TPS warping on the vertex coordinates of the target mesh. Left: warping a cube; Right: warping a person with clothes.}
    \label{fig:aug_3dtps}
\end{figure}

An example of the perturbed mesh is shown in \cref{fig:aug_3dtps}.

\subsection{Physical Color Calibration on the 3D Texture Map}

\begin{figure}[t]
\centering
\includegraphics[width=.8\linewidth]{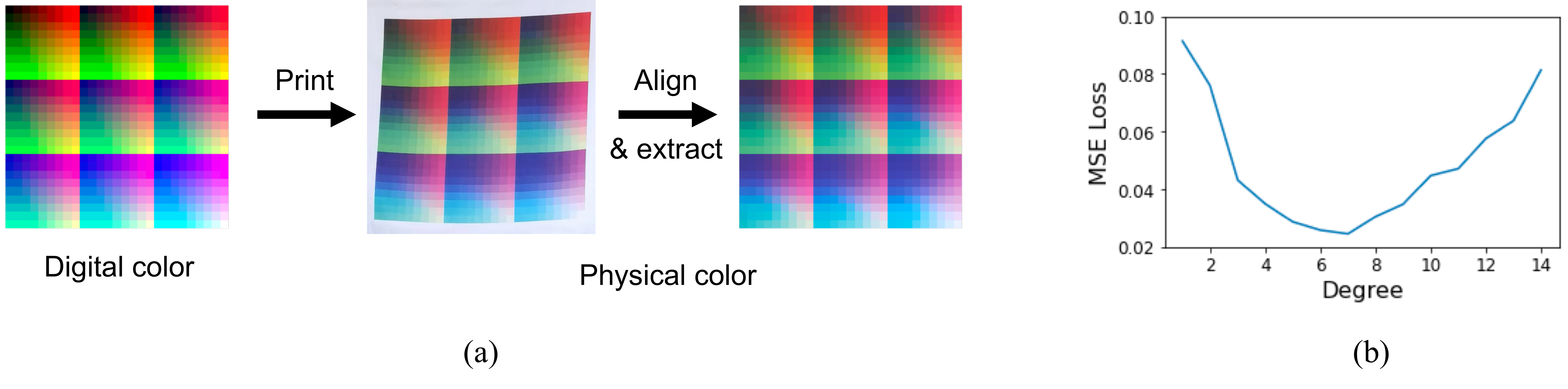}
\caption{Physical color calibration. (a) The pipeline of the calibration. The color palette consists of $729$ different colors. (b) The MSE loss of the fitting v.s. degree during validation. The MSE loss at each degree is averaged over $20$ partitions.}
\label{fig:color}
\end{figure}

As shown in \cref{fig:color}a, we first generate a color palette with $9\times9\times9=729$ different colors in the digital world. We then print it on a piece of cloth, take a picture for them, and extract the corresponding colors. We use a polynomial regression model with six degrees to fit the color transformation. Specifically, suppose the original digital color is $x=(R, G, B)$ and the final physical color is $y^*=(y_1^*, y_2^*, y_3^*)=(R^*, G^*, B^*)$. The regression model is
\begin{align}
    &y_i = \sum_{a_1,a_2,a_3}{w^{i}_{a_1,a_2,a_3}x_1^{a_1}x_2^{a_2}x_3^{a_3}}, i = 1,2,3,\\
    &a_1 + a_2 + a_3\leq \mathrm{d},
\end{align}
where $a_1, a_2, a_3$ and the degree $d$ are non-negative integers. A linear regression model is then applied to fit the polynomial features to $y^*$ with coefficients $w^{i}_{a_1,a_2,a_3}$. In order to choose a optimal degree $d$, we randomly divide the $729$ color pairs into training and validation sets, each containing $50\%$ of the pairs. We fit the model on the training set, and calculate the MSE loss on the validation set. As shown in \cref{fig:color}b, the MSE loss is the smallest when the degree is around $6$. Therefore, we choose $d=6$.

\section{Supplementary Results}
\subsection{Experimental Setup}

\vskip 5pt
\noindent\textbf{Datasets and Target Detectors.}
We collected $506$ background images in total from Google Search for 3D rendering. See \cref{sec:datasets} for some examples. We split them randomly into two sets for training and testing respectively, where the training set consists of $376$ images and the test set consists of $130$ images. 

We attacked three different detectors including YOLOv3~\cite{redmon2018yolov3}, Faster RCNN~\cite{ren2016faster}, and Deformable DETR~\cite{Zhu2021Deformable} under white box setting. We also attack other detectors including YOLOv2~\cite{redmon2017yolo9000}, Mask RCNN~\cite{he2017mask}, and DETR~\cite{carion2020end} to evaluate the transferability of the adversarial camouflage textures.

\vskip 5pt
\noindent\textbf{Evaluation Metric.} We extracted the bounding boxes from the output of the target detector on each input image and filtered out the boxes of which IoU scores with the ground truth box are lower than a specific IoU threshold $\tau_{\mathrm{IoU}}$. The threshold was set to $0.1$ in our paper except for Tab. 2 in the main text.
An image is regarded as an adversarial success example as long as the confidence scores of all the left boxes are lower than a confidence threshold $\tau_{\mathrm{conf}}=0.5$. The ASR is defined as the proportion of the adversarial success examples among all the test images. 

\vskip 5pt
\noindent\textbf{Implementation Details.}

We fixed the temperature $\tau$ in main text Eq. (\textcolor{red}{5}) to $0.3$ during training and use discrete sampling ($\tau\to0$) during evaluation and physical evaluation. We set the parameter $\lambda=0.7$ in main text Eq. (\textcolor{red}{6}). During training, we randomly perturbed the 3D models with the hyper-parameters $(\epsilon_r,\epsilon_t, \epsilon_{\mathrm{TPS}}) = (0.1, 50.0, 0.15)$. 
We optimized the parameters for $600$ epochs with Adam~\cite{kingma2014adam} optimizer with learning rate $0.001$ for the coordinates $b_{ij}$ in main text Eq. (\textcolor{red}{2}) and $0.01$ for the trainable Gumbel seed $u_i^{\mathrm{train}}$ in main text Eq. (\textcolor{red}{6}).

For digital evaluation, we rendered the mesh as we did for training, while using backgrounds sampled from the test set. We averaged the ASR over $37$ viewing angles ranging from $-180^{\circ}$ to $180^{\circ}$. Note that the 3D person model exactly faces the simulated camera when the viewing angle equals to $0^{\circ}$. 

For physical evaluation, we asked three actors to wear the adversarial clothes and turn a circle slowly in front of a camera, which was fixed at $1.55\;\mathrm{m}$ above the ground and $3.0\;\mathrm{m}$ distant from the actor unless otherwise specified. For each actor and each adversarial clothes, we recorded one video indoor and one outdoor. We extracted $32$ frames from each video and therefore collected $32\times3\times2=192$ for each clothes. We labeled the ground truth manually and evaluated the ASRs on these frames as we did in digital evaluation.


\subsection{Scene Dataset and Synthesized Images}
\label{sec:datasets}

\begin{figure}[t]
    \centering
      \begin{subfigure}{1.0\textwidth}
      \includegraphics[width=\textwidth]{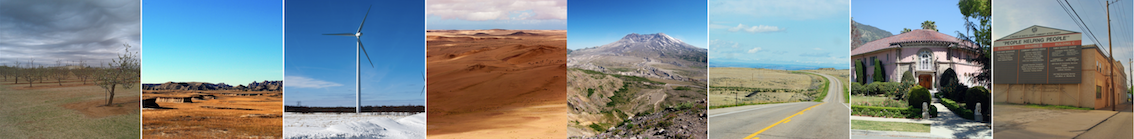}
      \caption{}
      \end{subfigure}
      \begin{subfigure}{1.0\textwidth}
      \includegraphics[width=\textwidth]{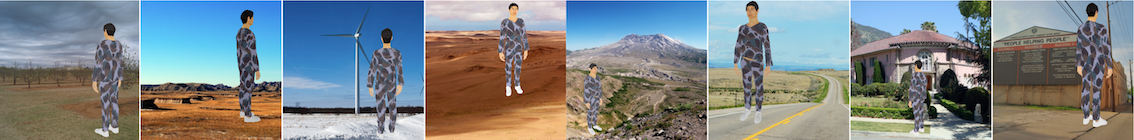}
      \caption{}
      \end{subfigure}
    \caption{Visualization of the dataset. (a) The background images. (b) The synthesized images.}
    \label{fig:backgrounds}
\end{figure}

See \cref{fig:backgrounds}a for examples of the background images in the scene dataset, and see \cref{fig:backgrounds}b for the synthesized images with rendered 3D person meshes as the foreground images. The 3D person meshes were rendered at different viewing angles, and were stuck onto the background images with random scales and positions.

\subsection{Visualization of the Bounding Boxes during Evaluation}

\begin{figure}[t]
    \centering
    \includegraphics[width=0.9\linewidth]{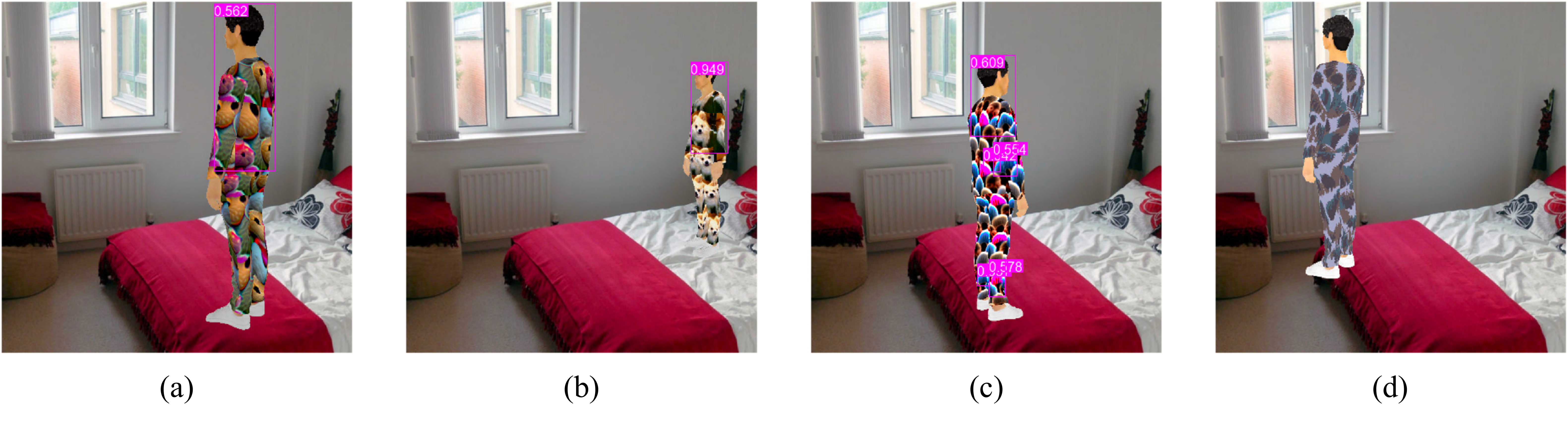}
    \caption{Visualization of the bounding boxes, with confidence threshold $0.5$. (a) AdvPatch~\cite{thys2019fooling}. (b) NatPatch~\cite{hu2021naturalistic}. (c) AdvTexture~\cite{hu2022adversarial}. (d) Ours.}
    \label{fig:iou_vis}
\end{figure}

See \cref{fig:iou_vis} for the examples of the detection results of different patterns. Previous methods were more likely to output some bounding boxes within the area of the foreground image, which had small but non-negligible IoU scores with the ground truth boxes. See Sec. 4.3 in the main text for the ASRs evaluated with different IoU threshold.

\subsection{Transfer Study in the Digital World}

\begin{figure}[t]
    \centering
    \includegraphics[width=0.7\linewidth]{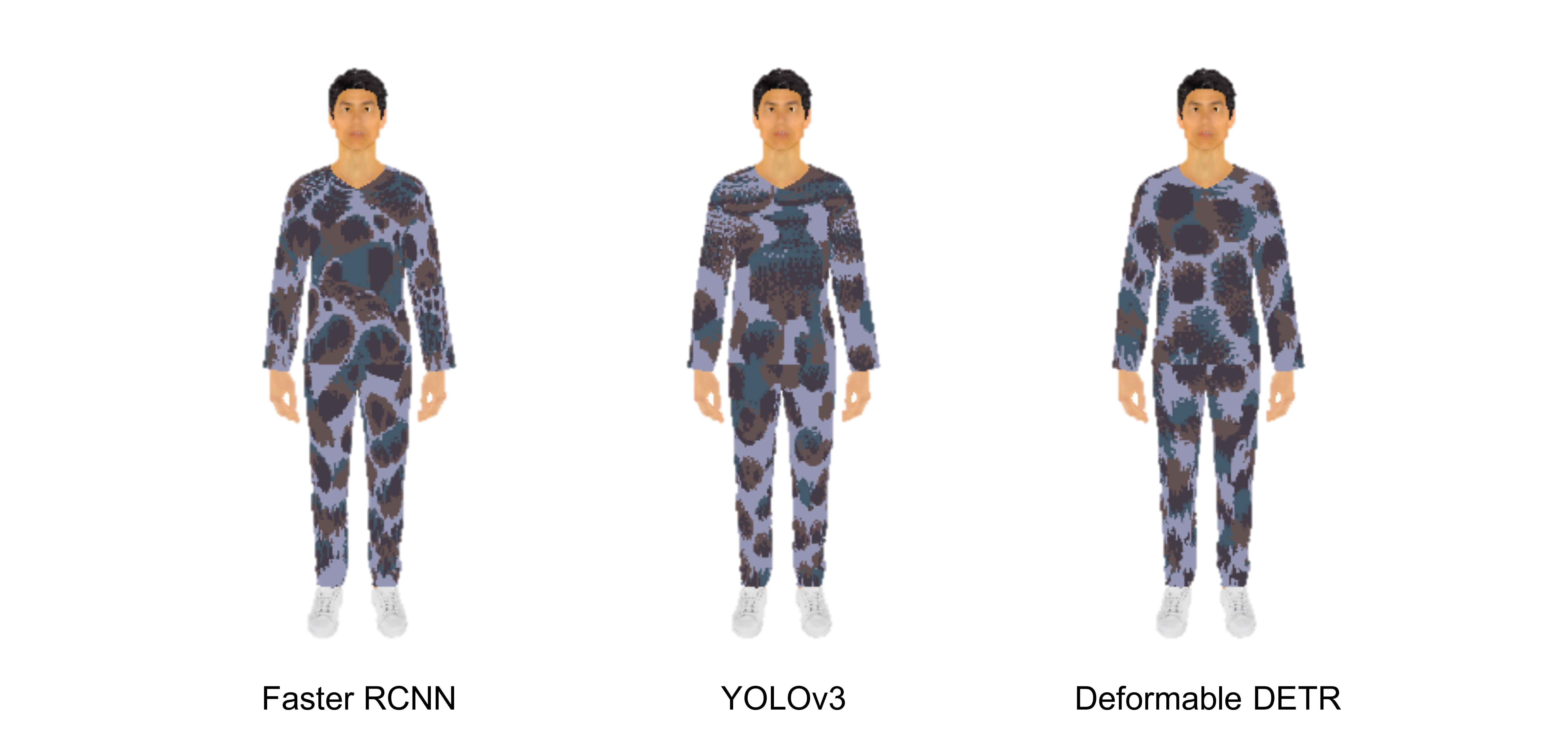}
    \caption{Visualization of the camouflage patterns against different detectors.}
    \label{fig:man_multi}
\end{figure}

We optimized camouflage patterns against different detectors including YOLOv3~\cite{redmon2018yolov3}, Faster RCNN~\cite{ren2016faster}, and Deformable DETR~\cite{Zhu2021Deformable}. \cref{fig:man_multi} shows the camouflage patterns against different detectors. See \cref{tab:transfer} for the ASRs of the transfer attacks in the digital world. The ASRs usually drop when the patterns were transferred to attack unseen detectors. In some cases, the ASRs still had high values (source Faster RCNN \& target Mask RCNN, $92.22\%$, source Deformable DETR \& targete Faster RCNN, $71.73\%$).

\begin{table}[t]
    \centering
    \small
    \begin{tabular}{lcccccc}
    \toprule
         \backslashbox{Source}{Target}&  Faster RCNN & YOLOv3 & Deformable DETR & Mask RCNN & YOLOv2 & DETR \\
    \midrule
        Random          & 0.85  & 3.31  & 5.76  & 0.50  & 2.10  & 4.95  \\
        Fast rcnn       & 99.36 & 28.21 & 65.11 & 92.22 & 18.05 & 36.05 \\
        YOLOv3          & 23.74 & 94.59 & 27.28 & 12.72 & 20.33 & 15.57 \\
        Deformable DETR & 71.73 & 23.26 & 88.77 & 43.99 & 16.38 & 48.19 \\

    \bottomrule
    \end{tabular}
    \caption{ASRs/\% of the transfer attacks in the digital world.}
    \label{tab:transfer}
\end{table}

\subsection{Parameter Sensitivity of $\lambda$}
\begin{figure}[t]
\centering
\includegraphics[width=.9\linewidth]{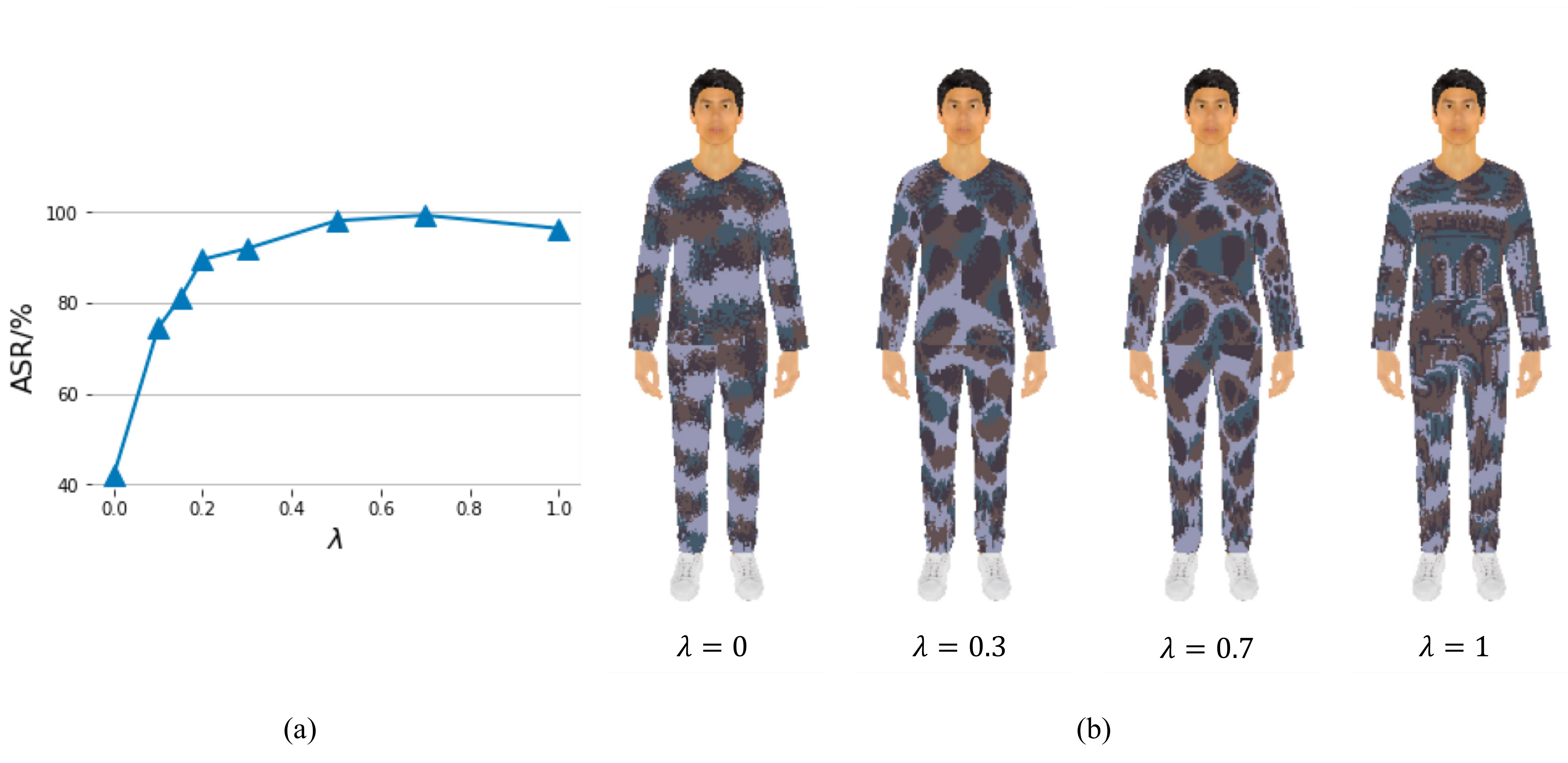}
\caption{Performance of different $\lambda$ values. (a) ASR v.s. $\lambda$. (b) Visualization of the pattern with different $\lambda$ values.}
\label{fig:lambdas}
\end{figure}

We optimized the AdvCaT patterns with different values of $\lambda$ in Eq.(6) in the main text, which controls the percentage of the trainable variable in the Gumbel seed. As shown in \cref{fig:lambdas}, small $\lambda$ resulted in poor adversarial effectiveness while large $\lambda$ resulted in strong adversarial effectiveness. Meanwhile, the AdvCaT pattern with small $\lambda$ was more like typical camouflage texture patterns, while the pattern with $\lambda=1$ looked somehow strange. Therefore, we chose $\lambda=0.7$ as a trade-off between the adversarial effectiveness and the appearance.

\subsection{Different Color Combinations of the AdvCaT Patterns}
\begin{figure}[t]
\centering
\includegraphics[width=.6\linewidth]{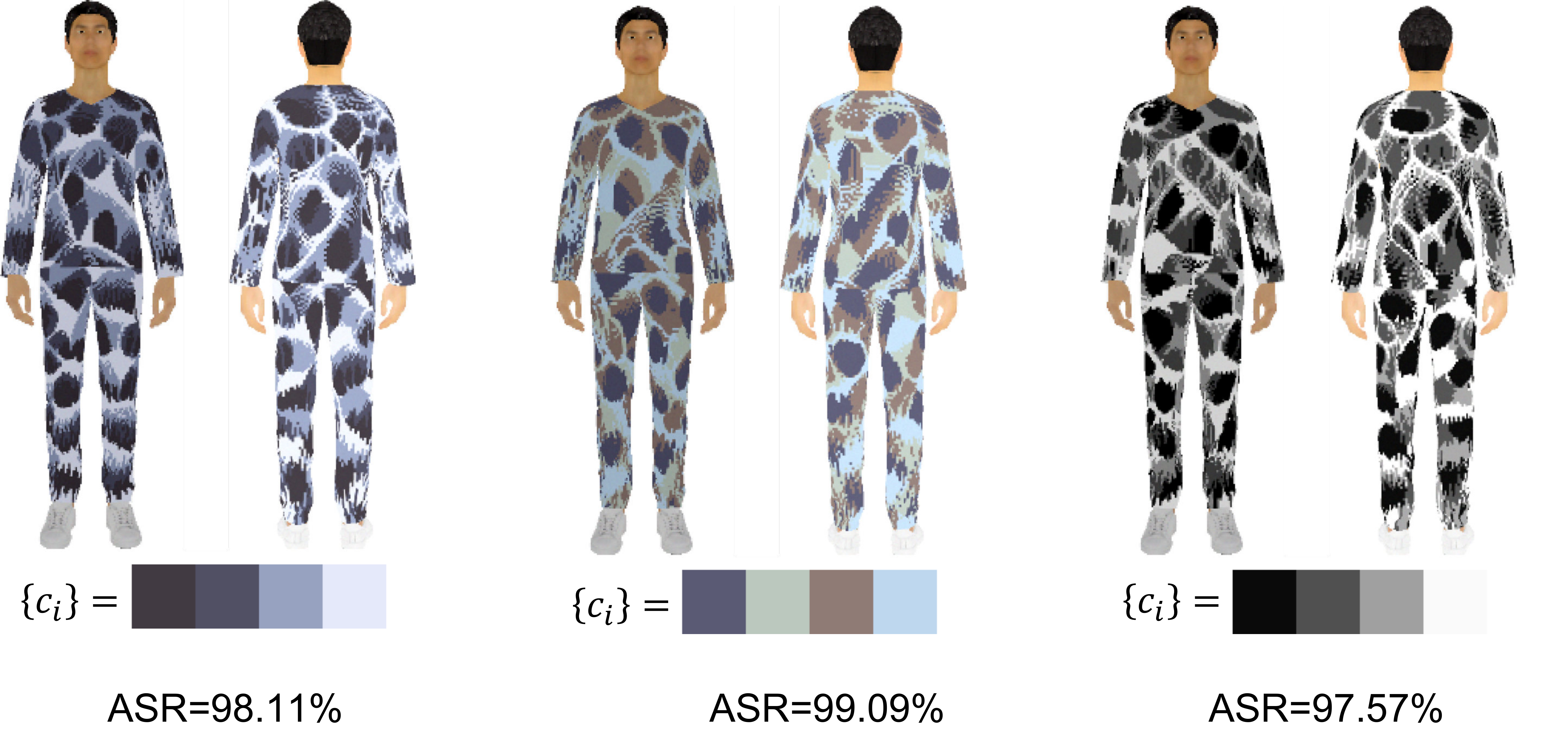}
\caption{ASRs of the patterns with different color combinations.}
\label{fig:different_colors}
\end{figure}

We used different color combinations to produce different styles of the AdvCaT patterns. See \cref{fig:different_colors} for the visualization. All of the AdvCaT patterns had high ASRs ($>97\%$) targeting Faster RCNN in the digital world.

\begin{table}[t]
    \centering
    \small
    \begin{tabular}{lll}
    \toprule
       & Indoor & Outdoor \\ \midrule
    Random & 0.00   & 0.00    \\
    AdvCaT & 86.46  & 85.42  \\ \bottomrule
    \end{tabular}
    \caption{ASRs/\% in different environments.}
    \label{tab:environment}
\end{table}

\begin{figure}[t]
\centering
\includegraphics[width=.35\linewidth]{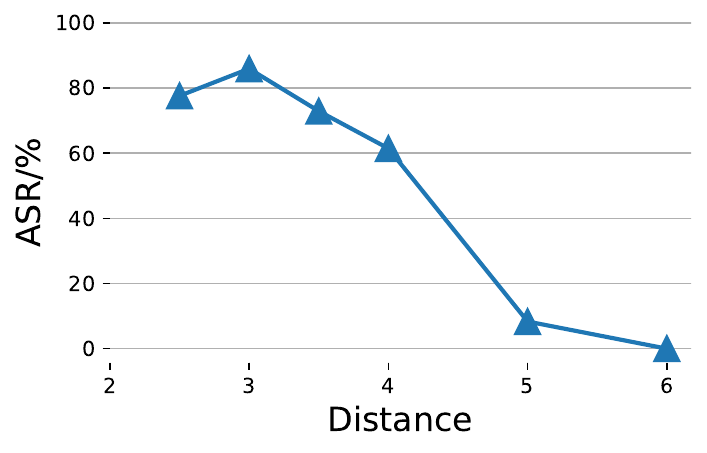}
\caption{ASRs v.s distances between the camera and actors.}
\label{fig:distance}
\end{figure}

\subsection{Attack in the Physical World under Different Physical Settings}
We evaluated the ASRs of the AdvCaT clothes under different physical settings. See \cref{tab:environment} for the ASRs of the adversarial  clothes in the indoor and outdoor scenes. See Sec. 4.1 in the main text for the collection of the physical test set. The ASRs of AdvCaT clothes were high (above $85.4\%$) both indoor and outdoor. In addition, \cref{fig:distance} shows the ASRs at different distances between the camera and actors. The ASRs stayed high (above $61.5\%$) when the distance was less than $4\;\mathrm{m}$.


\end{document}